\documentclass[10pt,lettersize,journal]{IEEEtran}
\usepackage{xcolor,pifont}

\usepackage[none]{hyphenat}
\usepackage{graphicx, caption}
\usepackage{hyperref}
\hypersetup{
     linkcolor    = red,
     colorlinks   = true,
     citecolor    = blue
}
\usepackage{tabulary, booktabs}
\usepackage{tabularx}
\usepackage{titlesec}
\usepackage{tikz}
\usepackage{amsmath}
\usepackage{amsfonts}
\usepackage{amssymb}
\usepackage{amsxtra}
\usepackage{mathrsfs}
\usepackage[ruled,linesnumbered]{algorithm2e}
\usepackage{multirow}
\usepackage{epstopdf}
\usepackage[export]{adjustbox}
\usepackage{balance}

\usepackage[mathscr]{}
\usepackage{todonotes}
\usepackage{xcolor}
\usepackage{tablefootnote}

\usepackage{enumitem}
\usepackage{braket}
\usepackage{color, colortbl}
\newcommand*\colourcheck[1]{%
  \expandafter\newcommand\csname #1check\endcsname{\textcolor{#1}{\ding{52}}}%
}
\newcommand*\colourcross[1]{%
  \expandafter\newcommand\csname #1cross\endcsname{\textcolor{#1}{\ding{55}}}%
}
\usepackage{dsfont}

\colourcheck{green}
\colourcross{red}
\newcommand{\myrowcolour}{\rowcolor[gray]{0.925}}
\newcommand{\name}{{HopTrack}\xspace}
\newif\ifdraft
\draftfalse
\newcommand{\xiang}[1]{\ifdraft{\textcolor{purple}{XL: #1}}\else{#1}\fi}

\author{\IEEEauthorblockN{ Xiang Li,
Cheng Chen,
Yuan-Yao Lou,
Mustafa Abdallah,
Kwang Taik Kim,
Saurabh Bagchi}\\
\IEEEauthorblockA{Purdue University \\
Email: \{li2068, chen4384, lou45, abdalla0, kimkt, sbagchi\}@purdue.edu}}

\begin{document}

\title{\name: A Real-time Multi-Object Tracking System for Embedded Devices}

%\author{IEEE Publication Technology,~\IEEEmembership{Staff,~IEEE,}
        % <-this % stops a space
%\thanks{This paper was produced by the IEEE Publication Technology Group. They are in Piscataway, NJ.}% <-this % stops a space
%\thanks{Manuscript received April 19, 2021; revised August 16, 2021.}}



\maketitle
\begin{tikzpicture}[remember picture,overlay]
\node[scale=0.8, color=gray!85, align=center, text width=15cm, anchor=north] at ([yshift=-1cm]current page.north) {
    \parbox{15cm}{\centering
        \footnotesize
        This work has been submitted to the IEEE for possible publication. \\
        Copyright may be transferred without notice, after which this version may no longer be accessible.}
};
\end{tikzpicture}
\begin{abstract}
Multi-Object Tracking (MOT) poses significant challenges in computer vision.
Despite its wide application in robotics, autonomous driving, and smart manufacturing, there is limited literature addressing the specific challenges of running MOT on embedded devices. 
State-of-the-art MOT trackers designed for high-end GPUs often experience low processing rates (\textless11fps) when deployed on embedded devices. Existing MOT frameworks for embedded devices proposed strategies such as fusing the detector model with the feature embedding model to reduce inference latency or combining different trackers to improve tracking accuracy, but tend to compromise one for the other. This paper introduces \name, a real-time multi-object tracking system tailored for embedded devices. Our system employs a novel discretized static and dynamic matching approach along with an innovative content-aware dynamic sampling technique to enhance tracking accuracy while meeting the real-time requirement. Compared with the best high-end GPU modified baseline Byte (Embed) and the best existing baseline on embedded devices MobileNet-JDE, \name achieves a processing speed of up to 39.29 fps on NVIDIA AGX Xavier with a multi-object tracking accuracy (MOTA) of up to 63.12\% on the MOT16 benchmark, outperforming both counterparts by 2.15\% and 4.82\%, respectively. Additionally, the accuracy improvement is coupled with the reduction in energy consumption (20.8\%), power (5\%), and memory usage (8\%), which are crucial resources on embedded devices. \name is also detector agnostic allowing the flexibility of plug-and-play. %Our implementation can be found at \url{https://anonymous.4open.science/r/HopTrack}
\end{abstract}


\section{Introduction} \label{sec: introduction}
 
Multi-Object Tracking (MOT) aims to detect and track multiple objects in video frames while preserving each object's unique identity across the frame sequence. This is usually accomplished by first running a detection model on a sequence of frames to identify objects, followed by a data association algorithm to link the same objects across frames.

\begin{figure}[t]
    \centering
\includegraphics[width=1\columnwidth]{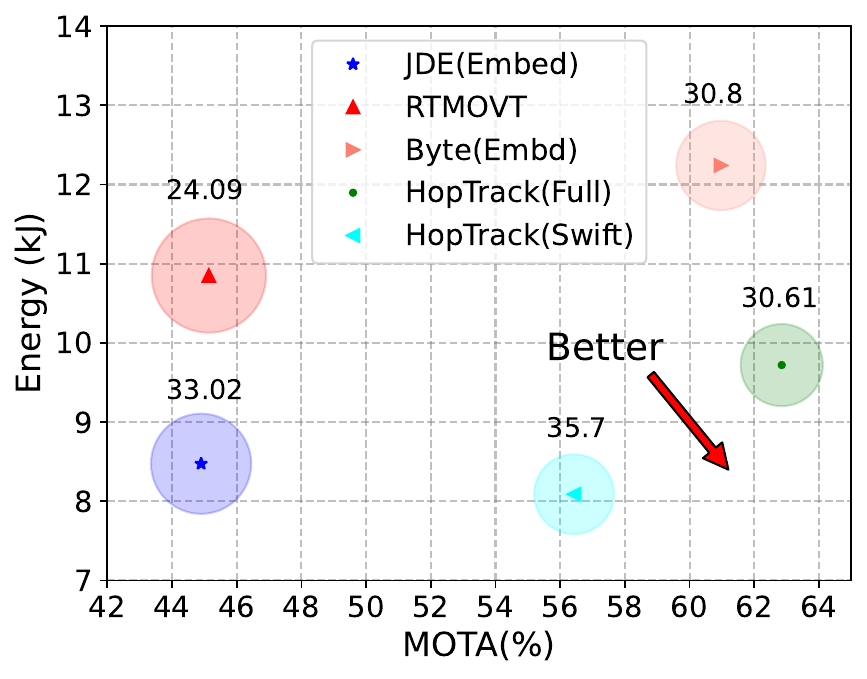}
\vspace{-8 mm}
\caption{Performance comparison of \name with baselines on embedded devices on MOT16. Circle size indicates the memory usage, while the number above the circle represents processing rate in fps.}
\vspace{-4 mm}
\label{fig:perfcomp}
\end{figure}

The challenge of MOT is two-fold. First, there can be drastic variations in the status and location of objects between frames, primarily caused by a low capture rate or algorithms that selectively process frames~\cite{sort,deepsort1,deepsort2}. Second, there is the issue of occlusion\footnotemark{} among objects in crowded scenes~\cite{occu_aware,occu_lit}.
\footnotetext[1]{Occlusion is a common term in computer vision and means that one object is partially or fully hiding one or more other objects in the same frame.}
The association between the same object across frames is typically achieved in two ways. The first approach uses a constant velocity model to predict the location of each object in frames following the detection frame~\cite{sort, zhang2022bytetrack}.
Then, it constructs a cost matrix based on the intersection over union (IoUs) of the actual detection bounding box in the subsequent frame and the predicted detection bounding box (i.e., complete overlap means the cost is 0, no overlap means the cost is 1).
The object association across frames is formulated as a linear assignment problem, which aims to minimize the cost by associating the detected bounding box with the predicted bounding box that has the highest IoU.
\begin{table*}[ht]
\resizebox{\linewidth}{!}{
\begin{tabular}{lllclccc}
\toprule[.2em]
\multicolumn{1}{c}{\textbf{Framework}}     & \multicolumn{1}{c}{\textbf{Testing  Device}} & \begin{tabular}[c]{@{}c@{}} \textbf{Benchmark}\\  \textbf{Dataset}\end{tabular}  & \multicolumn{1}{c}{\textbf{Real Time}\footnotemark{}} & \multicolumn{1}{c}{\textbf{FPS}\footnotemark{}(AGX fps)} &  \begin{tabular}[c]{@{}c@{}} \textbf{Open} \\  \textbf{Source} \end{tabular} & \begin{tabular}[c]{@{}c@{}} \textbf{Tracker Requires} \\  \textbf{Training} \end{tabular}  & \begin{tabular}[c]{@{}c@{}} \textbf{Detector} \\ \textbf{Independent} \end{tabular} \\
\midrule
SORT~\cite{sort}          & Intel i7 @ 2.5GHz     & MOT15 &       \greencheck   &   260 (association only)         &   \greencheck     &    \redcross    & \greencheck    \\
\myrowcolour
DeepSort~\cite{deepsort1,deepsort2}    & NVIDIA GTX 1050     & MOT16      &  \redcross   &   13.8 (5.9)          &   \greencheck     &    \greencheck  & \redcross      \\
% \hline
JDE~\cite{jde}           & NVIDIA Titan xp     & MOT16      &  \redcross   &   18.8 (9.08)        &   \greencheck     &    \greencheck  & \redcross      \\
\myrowcolour % \hline
StrongSort~\cite{du2022strongsort}           & Tesla V100    & MOT17/20   &  \redcross   &   7.4 (3.2)          &   \greencheck     &    \greencheck  & \redcross      \\
% \hline
ByteTrack~\cite{zhang2022bytetrack}     & Tesla V100     & MOT16/17/20 & \redcross   &   29.6 (10.11)         &   \greencheck     &    \redcross    & \greencheck    \\
\myrowcolour % \hline
OCSort~\cite{ocsort}     & NVIDIA RTX 2080Ti  & MOT17/20 & \redcross   &   28 (10.72)         &   \greencheck     &    \redcross    & \greencheck    \\
% \hline
RTMOVT~\cite{rtmot}        & Jetson TX2   & MOT16      &  \redcross   &   30 (24.1)           &   \redcross       &    \greencheck  & \redcross      \\
\myrowcolour % \hline
MobileNet-JDE~\cite{mobilenetjde} & Jetson AGX Xavier  & MOT16      &  \redcross   &   4.0 - 12.6   &   \redcross       &    \greencheck  & \redcross      \\
% \hline
REMOT ~\cite{remot}        & Jetson Xavier NX   & MOT16/17   &  \greencheck   &   58 - 81      &   \redcross       &    \greencheck  & \redcross      \\
\myrowcolour % \hline
\name (\textbf{Ours})& Jetson AGX Xavier   & MOT16/17/20   &  \greencheck   &   30.61  &   \greencheck     &    \redcross    & \greencheck    \\
\bottomrule
\end{tabular}
}
\vspace{0.5 mm}
    \caption{Comparison of \name and other MOT methods on embedded devices as well as high-end GPUs.} %The figures in parentheses represent our execution on the AGX Xavier platform.}
\label{tbl:all_com}
\vspace{-4mm}
\end{table*}

An alternative approach involves training a feature extractor model (embedding) that extracts deep features from the objects and uses those deep features to perform association through similarity comparison between objects across two frames~\cite{deepsort1,deepsort2}. 
Recent advancements involve the fusion of detection and embedding models to produce a joint detection and embedding (\textbf{JDE}) 
model to reduce the latency~\cite{jde,du2022strongsort,zhang2021fairmot}. 

However, these tracking methods predominantly rely on high-end GPUs. On the other hand, there are a growing number of applications, such as autonomous driving~\cite{auto1, auto2}, smart city surveillance~\cite{smartcity1,smartcity2}, and multi-robot collaboration in manufacturing~\cite{smart_manu, smart_manu_2}, where an accurate and fast MOT is needed but a high-end GPU is not practical due to physical, cost, and design constraints. Offloading computation to edge server or cloud~\cite{ibdash,coedge,distream, mtec} is a complementary approach as it can still benefit from more efficient local processing, which we provide. Further, offloading requires stable network connections, which are not always available in our target environments.

Designing an MOT system on embedded devices is challenging, because it is a resource-intensive, time-sensitive task, and the resources such as GPU power and memory are limited on these devices. 
Existing works such as REMOT~\cite{remot}, MobileNet-JDE~\cite{mobilenetjde}, and RTMOVT~\cite{rtmot} have attempted to address these challenges by exploiting the latency-friendly JDE architecture and performing detection on keyframes only, with tracking on the rest. 
However, these frameworks struggle with delivering high-quality results while meeting real-time processing needs. For example, MobileNet-JDE~\cite{mobilenetjde} operates at just 13 fps and RTMOVT~\cite{rtmot} achieves only 45\% tracking accuracy on the MOT16 test dataset.

\footnotetext[2]{The general definition of real-time processing rate is 24 frames per second (fps) as the typical video hardware capture rate ranges from 24 to 30 fps. In this work, we refer to a processing rate of 24-30 fps on NVIDIA Jetson AGX Xavier as near real-time and $\ge 30$ fps as real-time. The reported fps calculation includes both detection and association latency}

\noindent {\bf Our solution: \name}. In this paper, we present a real-time, multi-object tracking system, \name, specifically designed for embedded devices. 
\name brings three innovations to solve the problem. First, it dynamically samples the video frames (Section \ref{subsec: dynamic_sampling}) for detection based on the video content characteristics, e.g., complex scenes with plenty of objects and occlusions. 
Then, it employs two different data association strategies (Section \ref{subsec:data_association}), \textit{Hop Fuse} and \textit{Hop Update}, for fusing the detection results with existing track results and correcting tracking errors. 
\name uses innovative discretized static and dynamic matching techniques to analyze simple appearance features, such as pixel intensity distribution of different channels, and a trajectory-based data association method (Section \ref{subsec: traj_finding}) that can be computed efficiently on the CPU on {\em every} frame (Section \ref{subsec: image_quantization})  to achieve real-time, high-quality MOT.

\footnotetext[3]{For fair comparison, we downloaded all baseline frameworks where code was available or reimplemented them and then ran experiments on NVIDIA Jetson AGX Xavier. Only for REMOT, we used their reported metric values and used these to calculate the MOTA metric.}

Table~\ref{tbl:all_com} shows a comparative analysis of existing frameworks on both high-end GPUs and embedded devices (Jetson AGX). Figure~\ref{fig:perfcomp} illustrates \name's balanced performance in accuracy, processing rate, energy and memory usage comparing with baseline frameworks. Figure~\ref{fig: mota_imp} highlights the gradual accuracy improvement on embedded devices over the years, emphasizing the need for further exploration in this area. 
% SB (3/15/24): I do not see how we show this. Below in the contribution we have to claim that even the small improvement in accuracy that we achieve is significant in this problem context. 
We summarize our main contributions below.

\begin{enumerate}[leftmargin=*]
    \item We introduce \name, a real-time multi-object tracking framework for embedded device that achieves 63.12\% MOTA at around 30 fps on embedded device.
    \item We propose a dynamic and content-aware sampling algorithm that adjusts the running frequency of the detection algorithm.   
    \item We present a two-stage tracking heuristic called \textit{Hop Fuse} and \textit{Hop Update}, 
    which achieves an average processing speed of 30.61 fps and an average MOTA of 62.91\% on MOT16, 63.18\% on MOT17 and 45.6\% on MOT20 datasets. 
    \item We release our source code and models for the community to access and build on it. 
    We better all existing solutions for embedded devices in the accuracy or the processing speed (or both). 
\end{enumerate}

Our evaluation across multiple datasets (MOT16, MOT17, MOT20, KITTI), on a representative embedded device (NVIDIA AGX Xavier) brings out the following insights: (i) \name betters the state-of-the-art in accuracy, while maintaining real-time tracking (anything above 24 fps), with the closest competitor being Byte(Embed)~\cite{zhang2022bytetrack}; (ii) Reaching this involves a subtle interplay between detection and tracking on different frames and our microbenchmarks bring out that estimating trajectories of objects of different speeds is supremely important; (iii) It is an important advantage if processing can be largely on the CPU and in parallel; (iv) One has to carefully consider the power and the execution time to determine if a MOT solution is suitable for an embedded platform --- \name achieves the state-of-the-art in memory, power, and energy consumption. \xiang{We recognize the rapid pace of hardware development; however, these advancements are orthogonal to our research. For example, the Jetson Orin Nano (March 2023) offers 20 and 40 TOPS versions, comparable to the Xavier AGX's 32 TOPS but at a quarter of the cost and 2.7 times smaller. Our framework remains a more affordable, space-efficient solution as new hardware becomes available.}

The rest of the paper is organized as follows. In Section \ref{sec:problem_statement}, we provide a problem statement and discuss the challenges that motivate the development of \name. Section \ref{sec: framework} describes \name's framework in detail, including the algorithms of each component. In Section \ref{sec:experiment}, we present the experimental results, which demonstrate the effectiveness of \name in various settings on different benchmarks. Section \ref{sec:related_work} presents a comprehensive review of related work. Section \ref{sec:discussion} discusses the implications of our findings, and their potential applications in different domains, and outlines future directions. Section \ref{sec:conclusion} offers concluding remarks.

\section{Problem Statement and Key Challenges} \label{sec:problem_statement}
We summarize the key challenges of a multi-object tracking system on embedded devices here.
\begin{itemize}[leftmargin=*]
    \item \textbf{Low computation capability}: Despite improvement in the computation capabilities of embedded devices, such as the NVIDIA Jetson platforms, their inference time is still multiple factors of that of high-end GPUs due to the limited compute power. For instance, consider YOLOX~\cite{ge2021yolox}, an advancement in the YOLO series~\cite{redmon2016you, redmon2017yolo9000, redmon2018yolov3, bochkovskiy2020yolov4, glenn_jocher_2022_yolov5, ge2021yolox, yolov7} of object detectors. The inference time for YOLOX-S is around 10 ms on a V100 GPU but expands to around 80 ms on a Jetson AGX Xavier. The fastest YOLOv7~\cite{yolov7} network has an inference time of 6 ms on V100 GPU, but 60-70 ms on Jetson AGX Xavier (depending on the complexity of the scene in the MOT datasets). In this paper, we adopt YOLOX~\cite{ge2021yolox} as our object detector under accuracy and latency consideration. \xiang{Advanced detectors, such as transformer-based models, require more computational resources, which are not suitable for our application.} However, to demonstrate the detector-agnostic nature of \name, we also show integration with YOLOv7 in one experiment (Section~\ref{subsec:detector_agnostic}). \xiang{Other optimization techniques such as model quantization and distillation can further improve the inference speed. Since our framework is detector agnostic, such models can be easily integrated into our design.}

    \item \textbf{Time-sensitivity}: The approximate consensus for accepting a tracking operation is real time is to be 24 fps~\cite{real_time_1,real_time_2,real_time_3,real_time_4}. Hence, existing approaches that perform tracking based on detection outcomes on every frame, such as ByteTrack~\cite{zhang2022bytetrack} and JDE~\cite{jde}, are impractical on embedded devices as running detection on one single frame takes around 60-80 ms. On the contrary, a typical tracking frame on Jetson Xavier AGX only takes around 5-20 ms (depending on the tracking algorithm and scene complexity), which makes them suitable for real-time processing. We modified ByteTrack and JDE separately to create baselines Byte(Embed) and JDE(Embed)(Section ~\ref{baseline}), where we sampled the detection frames at a predefined frequency so that they meet the real-time requirement. Existing frameworks also adapted such frame sampling techniques~\cite{rtmot}. However, their  sampling technique is for a fixed rate, independent of the characteristics of the video, and this can significantly reduce tracking performance as the sampled detection frame may not be representative. Another commonly used approach to improve speed is through detection model compression or such as in MobileNet-JDE~\cite{mobilenetjde}, but it usually comes at the cost of detection accuracy.
    \item \textbf{Object association}: Performing object detection on every frame in a video sequence simplifies the association across consecutive frames because the objects' states typically do not change significantly within a short period of time. However, when detection is applied to frames separated by multiple frames, the association process becomes challenging because the objects' states may have changed significantly. Moreover, even between consecutive frames, the occlusions among objects in complex scenes add difficulty to the association process as the framework needs to be able to suppress the track when the object tracked is being occluded and re-identified it when it is unoccluded. To address those problems, RTMOVT~\cite{rtmot} combined JDE-modified YOLOv3 with a Kalman filter and KCF tracker~\cite{henriques2014high} to boost the association accuracy. REMOT~\cite{remot} enhances the feature embedding model with an angular triplet loss to increase the re-identification accuracy. A very recent paper~\cite{li2023multi} claims to do multi-object tracking on IoT devices through novel object-aware embedding, which enables them to achieve accurate, lightweight association. However, their evaluation is largely done on desktop GPUs (GTX 1080Ti).
\end{itemize}
 We identify two fundamental reasons why existing frameworks fail to perform real-time MOT on embedded devices: {\em (i)} inflexible, i.e., content-unaware sampling strategies, the framework samples the video at a constant rate or keyframes, which might not capture the changing dynamics of the video; and {\em (ii)} the detector-dependent JDE architecture poses a computational bottleneck as the it relies on the detectors to extract embedding features and detector execution is expensive. Therefore, lightweight tracking and the heavy weight detection are coupled together and cannot be performed separately.
 
\section{\name Design} \label{sec: framework}
\begin{figure}[t]
    \centering
\includegraphics[width=0.9\columnwidth]{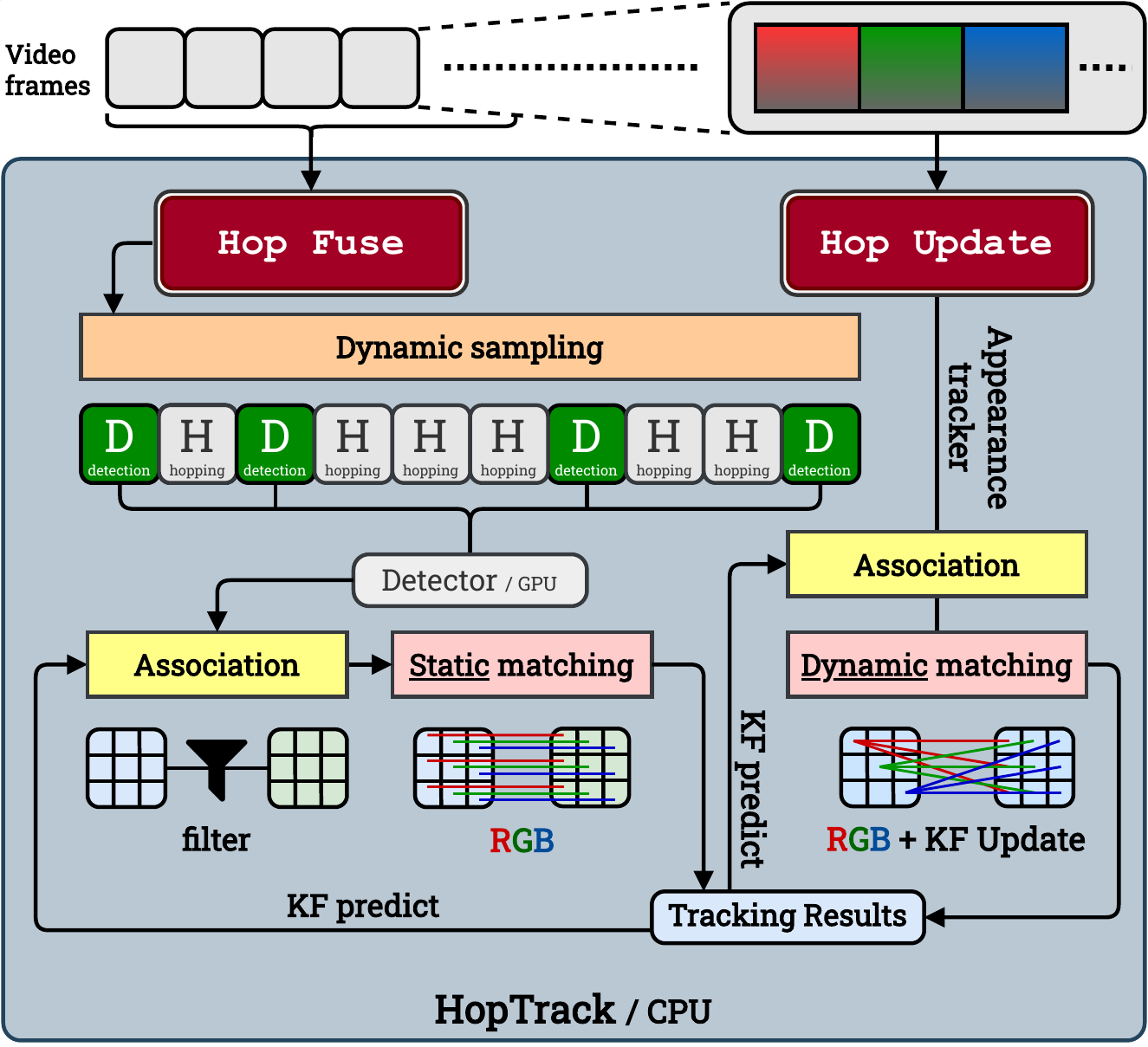}
\caption{System overview of \name. Hop Fuse associates active tracks with detections from dynamically sampled frames. Hop Update updates tracks' positions and suppresses inaccurate tracks.}
\label{fig:system_overview}
\vspace{-5 mm}
\end{figure}

Figure~\ref{fig:system_overview} shows the system overview of \name. The system addresses two primary challenges: dynamically sampling frames based on video content characteristics and performing data association across multiple frames.
To solve the first problem, we design a content-aware dynamic sampling algorithm (Section \ref{subsec: dynamic_sampling}) that adjusts the sampling rate based on the changing nature of the video content.
To solve the second problem, \name performs efficient data association (Section~\ref{subsec:data_association}), track adjustment on a per-frame basis using a trajectory-based track look-up method (Section \ref{subsec: traj_finding}), and shallow-feature-based, discretized static and dynamic matching (Section \ref{subsec: image_quantization} and Section \ref{subsec:dynamic-matching}).

\subsection{Dynamic Detection Frame Sampling}
\label{subsec: dynamic_sampling}
Current works frequently employ a key frame-based tracking strategy, where they run detection on key frames and run tracking algorithms such as OpticalFlow in between.
The key frames are extracted using a predefined interval or based on hints from the H.264 encoding~\cite{h264}. \xiang{To utilize H.264 encoding hints, I-frames typically serve as keyframes~\cite{videochef}, which are usually inserted every 80-120 frames, whereas in our dataset, they appear every 180-300 frames. At 30 fps, this results in a detection gap of 6-10 seconds, leading to missed details.} %where the I-frame refers to a complete frame (key frame).
This method is effective when the video is simple and no occlusions are present, but it falls short otherwise.

We design a content-aware, dynamic sampling algorithm to address this issue.
In this context, content refers to the number of objects of interest, as well as their respective sizes and positions inside the frame.
Our algorithm takes the detected bounding boxes from the detector as input, denoted as $b_{i} \in B$, and returns a sampling rate $\lambda$, with which detection will be run.
We calculate the centroids of each bounding box, denoted as $c_{i} \in C$.
Then, we perform our modified DBScan~\cite{ester1996density} algorithm, which can be described as follows:

$$N(c_{i}):\{c_{j}|d(c_{i},c_{j})\leq \epsilon_{1}, IoU(b_{i}, b_{j})>\epsilon_{2} \}.$$

Here we are calculating the neighbors of object $i$, which has centroid $c_i$, i.e., $N(c_i)$. For any other object $j$ in the frame, it will be included in the neighbor list if its euclidean distance from $c_{i}$ is below a threshold $\epsilon_1$, and the IoU between $b_{i}$ and $b_{j}$ is greater than a threshold $\epsilon_2$. The first condition is evident; the second condition ensures that the cluster is tight, as there are occlusions among objects in the cluster. 
This neighbor-finding process is repeated for $c_{i}$ and its neighbors, 
therefore $N(c_{i})$ keeps growing until no further neighbor can be included.
Eventually, if the total number of neighbors $N(c_{i})$ exceeds a threshold $M$, a cluster called $C_{i}$, which is grouped based on the neighbors of object $i$, is formed. 
Then, the system moves on to the next non-clustered object and uses that object as the center to start grouping new clusters.
In the end, we have a collection of clusters of close-by objects, denoted by ${C_{1}, C_{2}, ..., C_{n}}$.
The values used for $\epsilon_{1}$, $\epsilon_{2}$, and $M$ are empirically tuned to produce optimal performance.

\name dynamically adjusts the sampling rate\footnote{We use the term {\em sampling rate} to denote how often we have a detection frame in a cumulative set of detection and tracking frames. Thus, a sampling rate of 10 means we have 1 detection frame followed by 9 tracking frames.} based on the number of clusters as well as their density.
As the scene becomes packed with more clusters, \name algorithmically raises the sampling rate to acquire a more accurate estimation of each object's motion states to better predict the object's motion when they are occluded; when the scene is simpler, \name reduces the sampling rate.

\subsection{Data Association Algorithms} \label{subsec:data_association}
Motion blur, lighting, and occlusion can drastically reduce an object’s detection confidence  across the video sequence, resulting in association failure. Previous work often discard objects with low confidence~\cite{rtmot,rtmpt,mobilenetjde} or categorize them into low and high confidence categories before association. However, this strategy may fail when there is a long separation between detection frames, which are common in embedded devices.

We present a novel two-fold association method that significantly improves the association rate. The \textbf{Hop Fuse} algorithm is executed only when there is a new set of detection results available, and {\bf Hop Update} is performed on every hopping frame. In \textbf{Hop Fuse} \hyperref[alg:hop-fuse]{[Algorithm 1]}, the tracking pool $T$ is composed of the active tracks ($T_{active}$) from the previous frame and the lost tracks ($T_{lost}$). We define a track as active when it is not under occlusion or it can be detected by the detector when the object being tracked is partially occluded.
{\footnotesize
\begin{algorithm}[h]
    \SetAlgoLined
    \caption{Hop Fuse}
    \SetKwInOut{KwIn}{Input}
    \SetKwInOut{KwOut}{Output}
    \SetKwInOut{KwInit}{Initialization}
    \KwIn{Video sequence $V$; Object detector $Det$; Default sampling rate $\lambda$}
    \KwOut{Tracks $T_{active}$}
    \KwInit{$T \leftarrow \emptyset$, $T_{lost}\leftarrow \emptyset$, $sampling\_rate\leftarrow\lambda$}
    
    \For{frame id $f_{id}$, frame $f$ in $V$}{
        \# detection-track fuse / new track initialization frame\\
        \If{$f_{id} \% \lambda == 0$}{
            $T = T \cup T_{lost}$ \\
            $D_{id} = Det(f)$ \\
            $D_{tmp}, T_{active} = \emptyset$ \\
            \For{d in $D_{id}$}{
                \If{d.confidence $\geq \tau$}{
                    $D_{tmp} = D_{tmp} \cup \{d\}$ \\
                }
                $T_{pred} = \text{Kalman\_Filter\_Update}(T)$\\
                $T_{m}, T_{um}, D_{um} = \text{IoU\_matching}(T_{pred}, D_{tmp}, \phi_{1})$ \\
                $T_{active}.\text{add}(T_{m})$ \\
                $T_{m}, T_{um}, D_{um} = \text{IoU\_matching}(T_{um}, D_{um}, \phi_{2})$ \\
                $T_{active}.\text{add}(T_{m})$ \\
                $Traj = \text{Trajectory\_Finder}(T_{um})$ \\
                \For{$T_{i}$ in $Traj$}{
                    $D_{um}(j) = \text{Discretized\_Fix\_Match}(T_{i}, D_{um}, \psi_{1}, \psi_{2})$\\
                    $\text{Update}(D_{um}, T_{i}, Traj, T_{active})$\\
                }
                \For{$D_{i}$ in $D_{um}$}{
                    $T_{new} = \text{Create\_Track}(D_{i})$\\
                    $T_{active}.\text{add}(T_{new})$ \\
                }
            }
            $T_{lost}.\text{add}(Traj)$\\
            $\lambda = \text{Rate\_Adjust}(T_{active})$ running in a separate process\\ 
        }
    return $T_{active}$
    }
\label{alg:hop-fuse}
\end{algorithm}
}

Before data association, the detector of choice performs inference on the sampled frames $f$ to obtain the detection results $D_{id}$ and filters the result using a minimum acceptable confidence threshold ($\tau$). This filter prevents \name from erroneously tracking falsely detected objects.
Note that, instead of dividing the detection based on their confidence scores into two groups or setting a high confidence threshold, we empirically set a minimum confidence threshold $\tau$ of 0.4 as a lower bound to prevent erroneously tracking falsely detected objects. The Kalman filter is then applied to all tracks in $T$ to derive their predicted locations with bounding boxes ($T_{pred}$).

The first association is then performed based on IoU between $T_{pred}$ and filtered detections with a high threshold $\phi_{1}$, which primarily links stationary or minimally moving objects across several frames.
The matched tracks ($T_{m}$) are transferred from $T$ to $T_{active}$.
Then, a second round of IoU association is carried out with a lower threshold $\phi_{2}$ to link faster-moving objects with larger inter-frame displacements that were not matched previously ($T_{um}$).
Whenever a track and a new detection are successfully linked, the Kalman filter state of the original track is updated based on the new detection to enhance future movement prediction.

If there are still unmatched tracks, we proceed with trajectory discovery (Section \ref{subsec: traj_finding}) followed by discretized static matching (Section \ref{subsec: image_quantization}) to associate detections of objects that stray away from their original tracks.
For the rest of the unmatched detections, we consider them to be true new objects, create a new track for each, and assign them a unique ID. Any remaining unmatched tracks are marked as lost.
As for \textbf{Hop Update} \hyperref[alg:hop_update]{[Algorithm 2]}, unlike others that rely solely on either appearance tracker such as Optical Flow, MedianFlow, etc.~\cite{median_flow,KCF,boosting_tracker} or motion tracker like Kalman filter~\cite{sort, zhang2022bytetrack, jde}, we propose an appearance-motion combined tracking heuristic that leverages the strengths of both.
{\footnotesize
\begin{algorithm}[h]
    \SetAlgoLined
    \caption{Hop Update}
    \SetKwInOut{KwIn}{Input}
    \SetKwInOut{KwOut}{Output}
    \SetKwInOut{KwInit}{Initialization}
    \SetKwInOut{KwRe}{return}
    \KwIn{Video sequence $V$; Tracks from previous frame $T$}
    \KwOut{Actives Tracks $T_{active}$}
    \For{frame id $f_{id}$, frame $f$ in $V$}{
        \# track updates / suppress frame \\
        \If{$f_{id}$ \% $\lambda$ != 0 }{
            $T_{active} = \emptyset, T_{tmp} = \emptyset$ \\
            \For{$T_{i}$ in $T$}{
                \If{$T_{i}.new == True$}{
                    $T_{pred}=$ Appearance\_Tracker\_Update$(T_{i})$\\
                    $T_{i}.new = False$ \\
                    $T_{tmp}$.add($T_{pred}$)
                }\Else{
                    $T_{pred}$ = Kalman\_Filter\_Update($T_{i}$)\\
                    $T_{tmp}$.add($T_{pred}$$)$
                }
                 $T_{m}, T_{um}, P_{um}$ = IoU\_matching($T_{tmp}$, $T, \phi_{3}$) \\
                 $T_{active}$.add($T_{m}$) \\
                 $T_{m} =$ Discretized\_Dynamic\_Match$(T_{um}, P_{um}, \psi_{3}, \psi_{4})$\\
                 $T_{active}$.add($T_{m}$) \\
            }
        }
        $T_{lost}$.add($T_{um})$\\
        return $T_{active}$
    }
\label{alg:hop_update}
\end{algorithm}
}

In Hop Update, we use an appearance tracker (specifically, MedianFlow) for freshly produced tracks or those with reinitialized Kalman filter states ($T_{i}.new == True$) to obtain a predicted position $T_{pred}$ in the subsequent frame. The results of the appearance tracker are then used to adjust the object's Kalman filter state.
We empirically find that two updates from MedianFlow are sufficient to fine-tune the Kalman filter to produce reasonably accurate predictions.

\begin{figure*}

\begin{minipage}[t]{1.0\textwidth}
\begin{minipage}[t]{.32\textwidth}
\centering
  %\captionsetup{justification=centering}
\includegraphics[width=1.0\columnwidth]{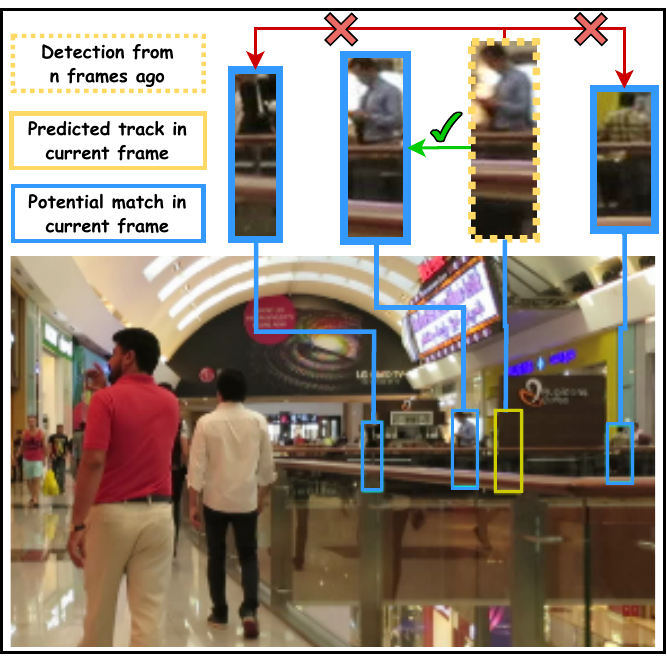}
     \caption[]{The yellow dashed box shows a prior detection, while the yellow box displays the current but incorrect tracking result. Blue boxes indicate candidate detections along the trajectory.}
    \label{fig: traj}
\end{minipage}\hfill
\begin{minipage}[t]{.32\textwidth}
\centering
 \captionsetup{justification=centering}
\includegraphics[width=1.0\columnwidth]{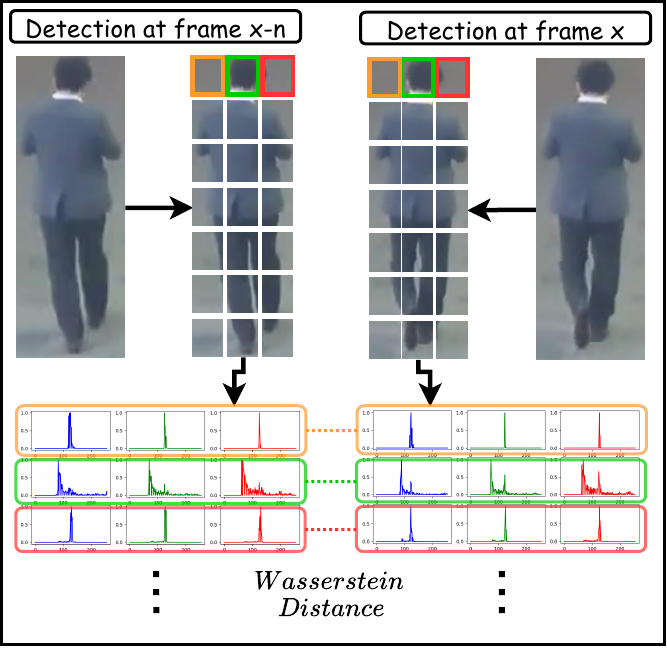}
     \caption{Discretized static matching calculates the intensity distribution for each channel in every image cell and computes the Wasserstein distance for corresponding pairs of cells.}
     \label{fig: static_image}
\end{minipage}
\hfill
\begin{minipage}[t]{.32\textwidth}
\centering
 \captionsetup{justification=centering}
\includegraphics[width=1.0\columnwidth]{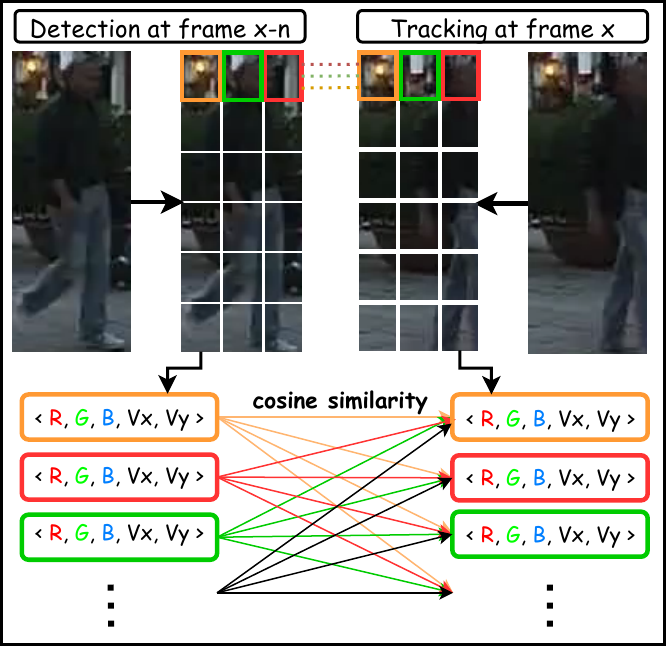}
      \caption{
     Dynamic discretized matching uses feature vectors \textless R, G, B, $\mathbf{v_{x}}$, $\mathbf{v_{y}}$\textgreater for each image cell, followed by cosine similarity calculation to assess if two tracks represent the same object.}
     \label{fig: dynamic_image}
\end{minipage} 
\end{minipage}
\vspace{-2 mm}
\end{figure*}

For objects that have been tracked for some time, we simply perform a Kalman filter update to obtain their predicted positions with bounding boxes in the subsequent frame. Then the identity association is performed between these predicted bounding boxes and the bounding boxes from the previous frame using an IOU matching followed by a discretized dynamic image match (Section \ref{subsec:dynamic-matching}).
To account for object occlusions, we perform discretized dynamic match on the predicted bounding boxes with the current frame's bounding boxes to intelligently suppress tracks when the object is under occlusion or when the Kalman filter state cannot accurately reflect the object's current state.
This method increases tracking accuracy by reducing missed predictions and by minimizing the likelihood that inaccurate tracks interfere with other tracks in future associations.
In the end, we add matched tracks to $T_{active}$, and for unmatched tracks, we either mark them as lost or remove them completely from the system if they have been lost for an extended time.
The active tracks are then sent into the next Hop Update or Hop Fuse to continue future tracking. 

\subsection{Trajectory-based Data Association} \label{subsec: traj_finding}
We propose a trajectory-based data association approach to improve the data association accuracy.
Unlike existing JDE approaches~\cite{jde,mobilenetjde,rtmot} that extract deep feature vectors and perform cosine similarity matching among all detections and tracks, we compute the predicted trajectory $Traj$ of unmatched tracks $T_{um}$ based on their Kalman filter states $\braket{x, y, a, h, v_{x}, v_{y}, v_{a}, v_{h}}$, which represents the centroids $(x, y)$, aspect ratio $(a)$, height $(h)$ of the objects and their respective changing rates ($v_{x}, v_{y}, v_{a}, v_{h}$).
Then, we project unmatched detections to $Traj$ and execute discretized static matching (Section \ref{subsec: image_quantization}) on those detections that are close to $Traj$.
The intuition behind this strategy is that if an object is moving quickly, then direction-wise, it cannot stray much from its initial path in a short amount of time, and vice versa. In addition, by eliminating detections that are located distant from the trajectory, we lower the likelihood of mismatch.

Figure \ref{fig: traj} illustrates our proposed approach.
The yellow box represents the object that we are interested in tracking, whereas the yellow box with dashes represents a prior detection several frames ago. Owing to various factors such as the erroneous state of the Kalman filter or the object's motion state change, the tracker deviates from the object of interest.
Three probable items (shown by blue boxes), which either lie on the trajectory of the object's original trajectory or the projection distance is close to the original trajectory, are presented as candidates; the rest of the objects in the scene are discarded during this round of trajectory matching. Next, discretized static matching is applied for association.

\subsection{Discretized Static Matching} \label{subsec: image_quantization}
The discretized static and dynamic matching is meant to use appearance features that can be extracted efficiently with the CPU, in order to associate objects with large inter-frame displacement across multiple frames and to suppress inaccurate tracks.
Static matching happens along with \textbf{Hop Fuse} only on the detection frames, and dynamic matching happens along with \textbf{Hop Update} on every hopping frame. 
In the JDE-based approach or the cascaded detection and embedding model approach, deep feature extraction requires intermediate layers' output from the detection model or a completely separate embedding model respectively.
Such feature extraction methods are costly and impractical on a per-frame basis on embedded devices.
Therefore, we propose combining CPU-efficient feature extraction and objects' motion states to perform object identity association.
During the \textbf{Hop Fuse} phase, the detector of choice (YOLOX-S) detects the object and marks the objects in the center of the bounding box with the bounding box enclosing the object as tightly as possible.
Then, a static discretized image matching is performed, as depicted in Figure ~\ref{fig: static_image}.
The left detection is from $n$ frames ago, where $n$ is determined by the current sampling rate $\lambda$, while the right detection is in the current frame.
For the static discretized detection matching, we discretize the detected object into $[M \times N]$ image cells and analyze each image cell individually.
By discretizing the image into image cells and performing pixel analysis, we can retrieve structural information from the image. Next, the Wasserstein distance is computed for each corresponding image cell's (normalized) pixel intensity distribution in the two detections. \xiang{Note that the 1D-Wasserstein distance calculation is performed on channel distributions and does not require image cells to be the same size.}
\begin{equation*}
\text{Match} = \boldsymbol{1} \left[ \left( \sum_{i,j} \boldsymbol{1} ({W}_{(i,j)} < \psi_{1}) \right) > \psi_{2} \right] \text{(i, j) are indices.}
\end{equation*}
Each pair of image cells is compared to an empirically set threshold $(\psi_{1})$. If greater than $\psi_{2}$ of the measured Wasserstein distances are below the threshold, then two detections are considered as the same object and we proceed with data association, and the motion state of the track is updated accordingly. We evaluate the sensitivity of our performance to choices of $(\psi_{1})$ and $(\psi_{2})$ in Section~\ref{sec:sensitivity}.
Importantly, the pixel intensity distribution and the 1-D Wasserstein distance calculations~\cite{wass_1, wass_2} for the image cells can be performed efficiently on the CPU and in parallel. The time complexity is $O(kz)$, where $k$ represents the number of detections that undergo the match process and $z=[M\times N]$ represents the number of discretized image cell pairs, which is tunable. Another commonly used feature that can be efficiently extracted with a CPU, and that could be used in static matching, is histogram of oriented gradients (HOG)~\cite{hog}. However, HOG features are subject to a fixed aspect ratio, which does not apply to our application scenario as a person's posture continuously changes.

\subsection{Discretized Dynamic Matching} \label{subsec:dynamic-matching}

The issue with static matching is that during the \textbf{Hop Update} phase, depending on the accuracy of the Kalman filter, the tracked objects might not be in the center of the bounding box or the bounding box might not be tight. 
Therefore, we propose a lightweight, dynamic discretized matching method to be run on each hopping frame, to check if the bounding boxes are accurately tracking the objects, and suppress tracks when occlusion happens.

As represented in Figure~\ref{fig: dynamic_image}, the bounding boxes of objects that potentially have the same identity across two frames are discretized into image cells $[M \times N]$ as in the static matching approach.
The actual detection result from the $n$-th previous frame is shown on the left, whereas the tracker-generated result is shown on the right.
These two bounding boxes are designated $B_{1}$ and $B_{2}$.

Since the position of the object may not be in the center of the bounding boxes, the previously utilized one-to-one image cell comparison of static matching is unreliable.
Instead, each image cell from the bounding box $B_{1}$ must be compared to each image cell of the potential match in $B_{2}$.
Thus, the number of Wasserstein distance calculations is $O(kn^{2})$, and they are performed on a frame-by-frame basis, this could result in a significant computation overhead~\cite{wass_overhead}. 
\begin{align}
\mbox{cosine}_{(i,j), (k,l)} \in \frac{B_{1}[F_{(i,j)}] \cdot B_{2}[F_{(k,l)}]}{\lVert B_{1}[F_{(i,j)}] \lVert * \lVert B_{2}[F_{(k,l)}] \lVert },
\label{equ:cos_sim}
\end{align}
Therefore, instead of calculating the distribution for each channel of each image cell pair and measuring the Wasserstein distance, we simply compute the average pixel intensity of each channel for each image cell and combine it with two of the Kalman filter states $v_{x},v_{y}$ from the tracked object to form a feature vector as follows: % Equation (\ref{equ:dynamic_vector}).
$F(P_{(i,j)}) = \ \braket{R_{(i,j)}, G_{(i,j)}, B_{(i,j)}, v_{x}, v_{y}}. $
Then, we calculate cosine similarity between each image cell from the bounding box $B_{1}$ and each image cell from the current frame's bounding box $B_{2}$ as Eq. (\ref{equ:cos_sim}):
where $B_{1}[F_{(i,j)}]$ and $B_{2}[F_{(k,l)}]$ denote the feature vectors for the image cells at position $(i,j)$ and $(k,l)$ in bounding boxes $B_{1}$ and $B_{2}$ for $i, k \in [1:M]$ and $j, l \in [1:N]$, respectively.

The matching problem is formulated as a linear assignment problem with a threshold $\psi_{3}$, where two image cells are considered a match when the cost (a metric that is the inverse of the cosine similarity) is less than $\psi_{3}$. %\kwang{Define threshold $\psi_{3}$.}
If the number of matched pairings exceeds a particular threshold $\psi_{4}$, we conclude that $B_{1}$ and $B_{2}$ track the same object. We then associate those two bounding boxes with the same identity and update the motion state of the track. We evaluate the sensitivity of our performance to choices of $(\psi_{3})$ and $(\psi_{4})$ in Section~\ref{sec:sensitivity}.

\section{Evaluation} \label{sec:experiment} 
In this section, we demonstrate that \name achieves the state-of-the-art accuracy (63.12\%) for embedded devices across multiple datasets, while simultaneously maintaining real-time processing speed (28.54 fps). Furthermore, we highlight the suitability of \name for embedded devices by showcasing its low power requirement (7.16W), small memory footprint (5.3G) and resource-efficient operation. Specifically, \name supports normal function with minimal 1 CPU core and an embedded GPU under up to 30\% contention, while maintaining a frame rate of 24 fps.

We evaluate \name under the private detector protocol on the MOT benchmark dataset~\cite{mot16_1,mot16_2, mot20}. We employed the well-tuned YOLOX-S detector following~\cite{zhang2022bytetrack}. %, which is pre-trained on the COCO dataset~\cite{COCO} and further trained on the MOT17 training sets, CrowdHuman~\cite{shao2018crowdhuman}, Cityperson~\cite{zhang2017citypersons} and ETHZ~\cite{ess2008mobile}.
The details of specific sequences in the MOT16, MOT17, and MOT20 test datasets are listed on the following websites~\cite{mot16dataset, mot17dataset, mot20dataset}. It comprised of video with different resolutions (480p, 1080p, non-standard) and different light conditions (day and night). All test results are carried out on the NVIDIA Jetson AGX Xavier embedded device, which is widely used in industry for applications such as robotics. \xiang{We emphasize that the solution is not optimized solely for a specific device. The only device-specific design choice in HopTrack pertains to the upper and lower bounds of the detection rate, which are determined by the detector’s inference time on that specific device and also impact other baselines.}

We performed an ablation study on each design choice meticulously while comparing with baseline frameworks. To assess the effectiveness of content-aware dynamically sampling, we designed variants of \name: \textbf{HopTrack(Acc)}, \textbf{HopTrack(Swift)}, and \textbf{HopTrack(Full)}.
HopTrack(Acc) samples detection frames at the highest constant frequency that will allow the framework run at the real-time requirement of 30 fps, while HopTrack(Swift) samples detection frames at a lower constant frequency that maintains tracking accuracy within 10\% degradation compared to HopTrack(Acc). HopTrack(Full) incorporates the content-aware sampling technique. We fix the upper and the lower bounds of the sampling to match that of HopTrack(Acc) and HopTrack(Swift). Then, HopTrack(Full) is free to vary the sampling rate based on the video content characteristics within this range.

We evaluated trajectory-based matching by conducting a side-by-side comparison with and without it enabled for each variants (Table~\ref{tbl:baseline_cmp}\&\ref{tbl:hopswift_traj}). We examined \name's performance under both CPU and GPU resource contention (Sec~\ref{subsubsec:rescontention}). To demonstrate the detector-agnostic design, we integrated \name with two different detectors to showcase its advantage over the baseline (Sec~\ref{subsec:detector_agnostic}).

\subsection{Metrics}
\textbf{Tracking Accuracy}: 
We use the CLEAR metrics~\cite{clear_mot}, including MOTA, IDF1~\cite{idf1}, FP (False Positive), FN (False Negative), IDSW (ID switch), HOTA (higher order tracking accuracy)~\cite{luiten2021hota} to evaluate different aspects of \name.
\begin{itemize}[leftmargin=*]
    \item{\textbf{MOTA($\uparrow$)}}: Multi-object tracking accuracy computed as $1 -(\mbox{FP}+\mbox{FN}+\mbox{IDSW})/\mbox{GT}$. GT represents ground truth.
    \item{\textbf{IDF1$(\uparrow)$}}~\cite{idf1}: Ratio of correctly identified detections over the average of ground truth and computed detections. It offers a single scale that balances identification precision and recall.  
    \item{\textbf{IDSW}$(\downarrow)$}: The number of identity switches that occur to the same object during its lifetime in the video sequence. (i.e. the tracker switches the identities of two objects, or the tracker lost track and reinitialized it with a new identity.)
    \item{\textbf{HOTA$(\uparrow)$}}~\cite{luiten2021hota}: This recently proposed metric is calculated as the average of the geometric mean of detection and association accuracy. 
\end{itemize}

\subsection{Baselines}
\label{baseline}   
    \begin{itemize}[leftmargin=*]
        \item{\textbf{RTMOVT~\cite{rtmot}}}: A real-time visual object tracking method that adopts the JDE approach to run on embedded GPU.
        \item{\textbf{MobileNet-JDE~\cite{mobilenetjde}}}: A JDE-based tracker for embedded devices with redesigned embedding head and anchor boxes by the original authors of MobilNet-JDE~\cite{mobilenetjde}.
        % SB (11/30/23): Original authors of JDE?
        % XL (11/30/23) : fixed
        % SB (6/29/23): Redesigned by us?
        % XL (6/29/23): Redesigned by the author
        \item{\textbf{REMOT~\cite{remot}}}: A resource-efficient tracker designed for embedded devices utilizes model compression techniques to speed up inference.
        \item{\textbf{Byte(Embed)~\cite{zhang2022bytetrack}}}: A modified version of the state-of-the-art tracker Bytetrack that can execute in real-time on embedded devices.
        \item{\textbf{JDE(Embed)~\cite{jde}}}: A modified version of JDE framework that can execute in real-time on embedded devices.
    \end{itemize}

\subsection{Evaluation Results} \label{subsec:eva}

% \begin{figure}[h]
%     \centering
% \includegraphics[width=0.9\columnwidth]{Figures/radarChart.pdf}
% \vspace{-3mm}
%      \caption{Comparison of \name and baselines in terms of MOTA, fps (frame rate), energy and power consumption, and memory usage. \name has a well-balanced performance in all 5 aspects.}
%      \label{fig:baseline_compare}
% \vspace{-2mm}
% \end{figure}
 
%\xiang{Figure \ref{fig:baseline_compare} shows the well-balanced performance of \name in with respect to tracking accuracy, frame rate, energy, power, and memory usage (power, energy and memory usage are labeled in reverse order). Figure \ref{fig:mota_improvement} shows the MOT performance improvements on embedded devices in the literature.}

\subsubsection{\textbf{Tracking Accuracy Evaluation}}~
Table \ref{tbl:baseline_cmp} details CLEAR metrics and processing rate comparison of \name and the baselines. 
In Table~\ref{tbl:baseline_cmp}, HopTrack(Acc) achieves the best MOTA score of 63.12\%, surpassing the best baseline Byte(Embed) by 2.15\%. For the IDF1 metric, HopTrack(Full) beats the best baseline by 2.45\%, and for the HOTA metric, we exceed the best baseline by 1.65\%. Despite REMOT~\cite{remot} having the lowest number of IDSW (791), there is a large trade-off between false negatives (82903) and the low IDSW. 
All frameworks, except MobileNet-JDE~\cite{mobilenetjde}, are capable of real-time execution.
\begin{table}[h]
\resizebox{\columnwidth}{!}{
\begin{tabular}{lccccccr}
\toprule[.2em]
& \multicolumn{1}{c}{\textbf{MOTA\%}} & \multicolumn{1}{c}{\textbf{IDF1\%}} & \multicolumn{1}{c}{\textbf{HOTA\%}} & \multicolumn{1}{c}{\textbf{IDSW}}  & \multicolumn{1}{c}{\textbf{FP}} & \multicolumn{1}{c}{\textbf{FN}} &\multicolumn{1}{c}{\textbf{FPS}}\\
\midrule
Byte(Embed)~\cite{zhang2022bytetrack}   & 60.97  & 58.38  & 48.70       & 2958 & 16232 & 51963 & 30.8\\
\myrowcolour
JDE(Embed)~\cite{jde}                   & 44.90  & 45.36  & 36.13       & 2474 & 17034  & 80949 & 33.02 \\
RTMOVT~\cite{rtmot}                     & 45.14  & 45.54  & 36.78       & 2886 & 18943 & 78195&  24.09\\
\myrowcolour
MobileNet-JDE~\cite{mobilenetjde}       & 58.30  & 48.00  & 41.00       & 3358 & 9420  & 63270 & 12.6\\
REMOT\footnotemark{} ~\cite{remot}      & 48.93  & 54.40  & ---         & \textbf{791}  & \textbf{9410}&  82903  & \textbf{58}\\
\hline
\myrowcolour
HopTrack(Full)                               & 62.91  & \textbf{60.83}  & \textbf{50.35}  & 2278 & 19063 & 46283 & 30.61\\
\hspace{6mm}    w/o trajectory          & 62.80  & 60.76  & 50.30      & 2331 &  19037 & 46464 & 34.18\\
\myrowcolour
HopTrack(Swift)                                & 56.43  & 57.50  & 47.50     & 2348 &25307 & 51927 & 35.7 \\
\hspace{6mm}    w/o trajectory          & 55.62  & 57.04  & 47.27     & 2452 & 25713 & 52760 & 39.29\\
\myrowcolour
HopTrack(Acc)                             &\textbf{63.12}& 60.70 & 50.26  & 2184 & 18898 & \textbf{46158} & 28.54\\
\hspace{6mm}    w/o trajectory          & 62.85  & 60.76  & 50.35     & 2383  & 18890 & 46470 & 31.89 \\
\bottomrule % \cline{1-11}
\end{tabular}
}
\caption{\name and baseline test results on MOT16 test dataset. \name and its variants achieve higher tracking accuracy compared to existing baselines.}
\label{tbl:baseline_cmp}
\end{table}
\footnotetext{We noticed that the MOTA score reported by REMOT does not correctly match the other metrics they reported in their paper. We attempted to contact the author but received no response. The MOTA value presented in this table for REMOT is numerically computed based on the FP, FN, and IDSW values reported in the original paper.}
\begin{table}[h]
\resizebox{\linewidth}{!}{
\begin{tabular}{ccccrrr}
\toprule[.2em]
\multicolumn{1}{c}{\textbf{Sequence}}      & \multicolumn{1}{c}{\textbf{MOTA\%}}  & \multicolumn{1}{c}{\textbf{IDF1\%}} & \multicolumn{1}{c}{\textbf{HOTA\%}}   & \multicolumn{1}{c}{\textbf{FP}}     & \multicolumn{1}{c}{\textbf{FN}}    & \multicolumn{1}{c}{\textbf{IDSW}} \\
\midrule
\textbf{HopTrack(Full)} \\
MOT16-01 & 54.35 & 60.42 & 48.26 & 516   & 2376  & 27   \\
\myrowcolour
MOT16-03 & 81.15 & 72.85 & 59.45 & 4163  & 15356 & 186  \\
MOT16-06 & 42.5 & 44.39 & 36.08 & 2142  & 3934  & 558  \\
\myrowcolour
MOT16-07 & 37.99 & 29.40 & 28.91 & 3331  & 6331  & 460  \\
MOT16-08 & 38.64 & 47.73 & 40.51 & 3396  & 6668  & 205  \\
\myrowcolour
MOT16-12 & 27.37 & 37.18 & 31.25 & 2564  & 3294  & 167  \\
MOT16-14 & 35.35 & 50.63 & 36.42 & 2951  & 8324  & 675 \\
\midrule %\midrule % \hline \hline
OVERALL  & \textbf{62.91} & \textbf{60.08} & \textbf{50.35} & 19063 & \textbf{46283} & \textbf{2278} \\
\textbf{BYTE(Embed)} \\
MOT16-01 & 52.02 & 57.44 & 46.08 & 320   & 2724  & 21   \\
\myrowcolour
MOT16-03 & 81.21 & 72.25 & 58.23 & 3127  & 16332 & 185  \\
MOT16-06 & 42.73 & 44.07 & 36.27 & 1714  & 4274  & 620  \\
\myrowcolour
MOT16-07 & 34.50 & 28.21 & 28.20 & 2996  & 7212  & 483  \\
MOT16-08 & 39.91 & 47.27 & 40.43 & 2670  & 7203  & 184  \\
\myrowcolour
MOT16-12 & 33.73 & 36.92 & 30.94 & 1745  & 3577  & 175  \\
MOT16-14 & 15.65 & 27.73 & 24.08 & 3660  & 10641  & 1290 \\
\midrule %\midrule
OVERALL  & 60.97 & 58.38 & 48.7 & \textbf{16232} & 51963 & 2958 \\
\bottomrule
\end{tabular}
}
\caption{Tracking results for HopTrack(Full) and Byte(Embed) on MOT16 dataset. {\em \small (The IDs of the sequences are not continuous; the missing ones are from the training set. The overall is not a simple average of the measures of all the sequences as the sequences have different lengths in terms of the number of frames and tracking objects.)}}
\label{tbl:mot16_dynamo}
\end{table}

Table \ref{tbl:mot16_dynamo} shows the comparison of the detailed testing results for each testing sequence in the MOT16 test dataset of HopTrack(Full) and Byte(Embed). We choose Byte(Embed) as it is the most competitive baseline considering the balance of accuracy and speed. We note that HopTrack(Full) outperforms Byte(Embed) in overall MOTA by 1.94\%. In particular, the improvement is striking, nearly 20\%, 
on the MOT16-14 sequence, which was captured from a moving bus. This significant performance improvement can be attributed to innovative trajectory-based matching. When objects move fast (as in the case of the moving vehicle), Byte(Embed) struggles with data association due to the low IoU between detection bounding boxes across multiple frames and the abrupt change in detection confidence. On the other hand, HopTrack(Full) continues tracking objects through trajectory-based finding and discretized matching, which allows for accurate identity association across frames, thus leading to superior performance compared to Byte(Embed).

\begin{table}[t]
  \resizebox{\linewidth}{!}{
  \begin{tabular}{ccccrrr}
  \toprule[.2em]
  \multicolumn{1}{c}{\textbf{Scheme}}      & \multicolumn{1}{c}{\textbf{MOTA\%}}  & \multicolumn{1}{c}{\textbf{IDF1\%}} & \multicolumn{1}{c}{\textbf{HOTA\%}}   & \multicolumn{1}{c}{\textbf{FP}}     & \multicolumn{1}{c}{\textbf{FN}}    & \multicolumn{1}{c}{\textbf{IDSW}} \\
  \midrule
HopTrack(Swift) & 56.43 & 57.50 & 47.50 & 25307  & 51927 & 2348  \\
\myrowcolour
- MOT16-14 & 28.35 & 43.80 & 32.30 & 3611   & 8989  & 659   \\
\midrule% \hline
No-Traj\footnotemark{} & 55.60 & 57.04 & 47.27 & 25713 & 52760 & 2452\\
\myrowcolour
- MOT16-14 & 21.23 & 37.42 & 28.39 & 4076  & 9728  & 755   \\
\bottomrule
\end{tabular}
}
\caption{HopTrack(Swift) with and without trajectory-based matching. Result of the fast moving video sequence MOT16-14 is highlighted.}
\label{tbl:hopswift_traj}
\vspace{-6 mm}
\end{table}
\footnotetext{HopTrack(Swift) with trajectory-based matching disabled.}

\begin{figure*}

\begin{minipage}[t]{1.0\textwidth}
\begin{minipage}[t]{.33\textwidth}
\centering
  \captionsetup{justification=centering}
\includegraphics[width=1.0\columnwidth]{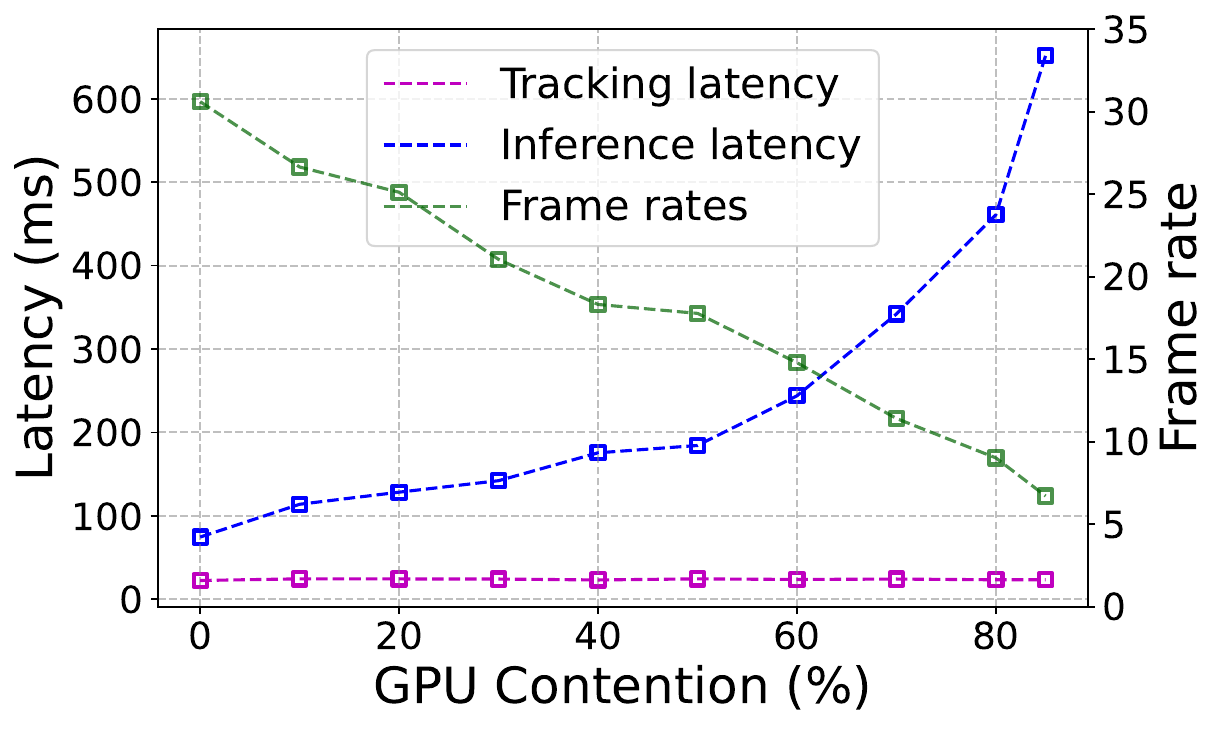}
    \vspace{-7mm}
     \caption{The performance of \name under different GPU contention levels. }
    \label{fig: gpu_stress}
\end{minipage}\hfill
\begin{minipage}[t]{.33\textwidth}
\centering
 \captionsetup{justification=centering}
\includegraphics[width=1.0\columnwidth]{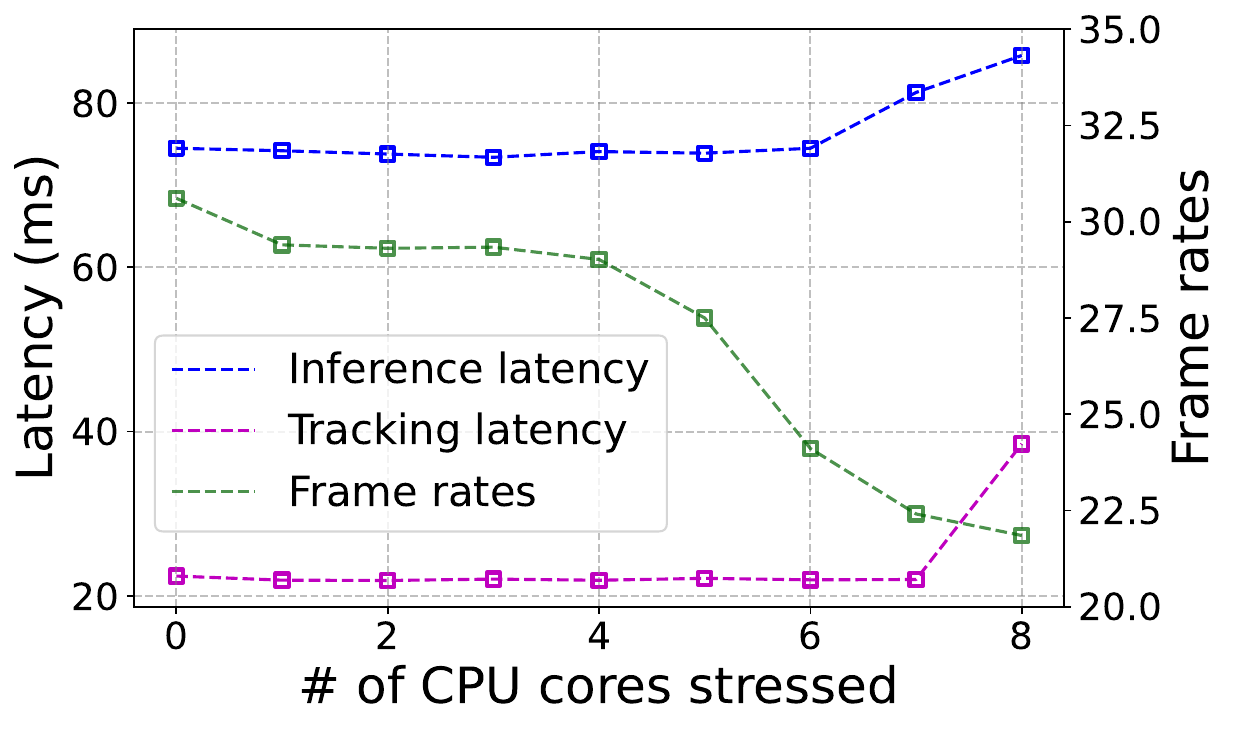}
\vspace{-7mm}
     \caption{The performance of \name as the number of stressed CPU cores increases.}
     \label{fig: cpu_stress}
\end{minipage}
\hfill
\begin{minipage}[t]{.33\textwidth}
\centering
 \captionsetup{justification=centering}
\includegraphics[width=1.0\columnwidth]{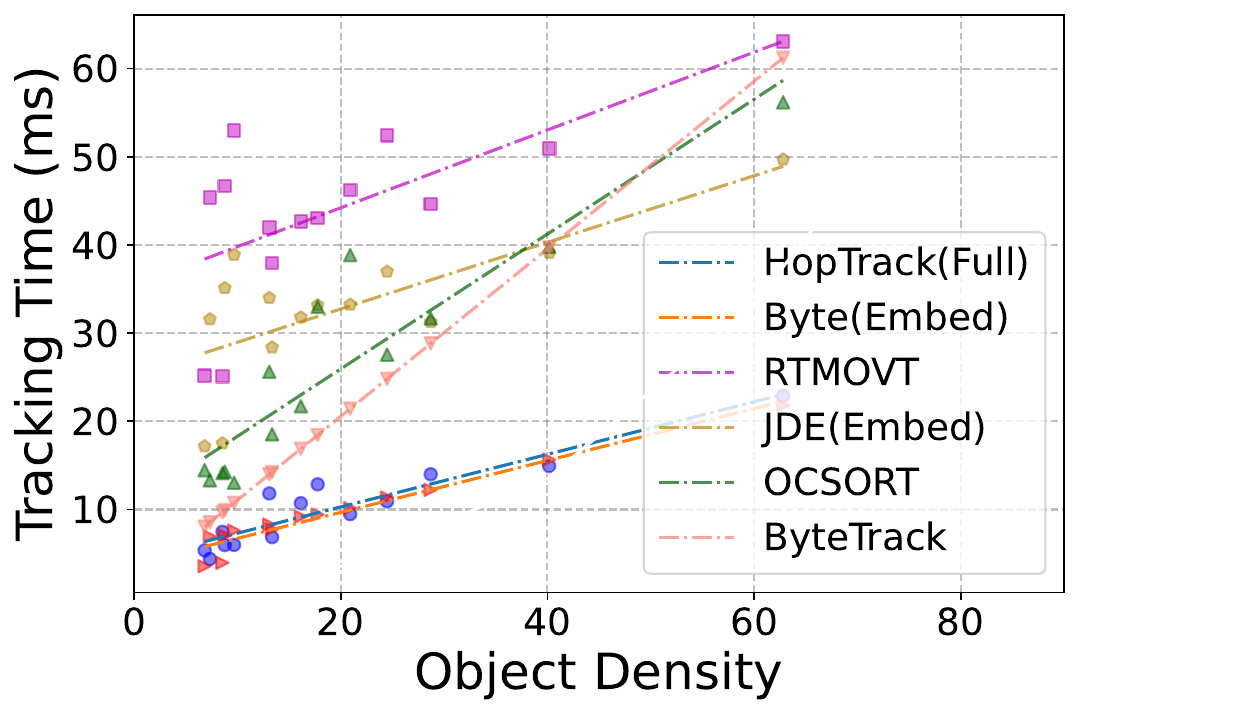}
\vspace{-7mm}
      \caption{
     \name tracking time is one of the least impacted by object density (MOT20 dataset).}
     \label{fig: tracking_time}
\end{minipage} 

\end{minipage}
\vspace{-3mm}
\end{figure*}

\begin{figure*}

\begin{minipage}[t]{1.0\textwidth}
\begin{minipage}[t]{.33\textwidth}
\centering
  \captionsetup{justification=centering}
\includegraphics[width=1.0\columnwidth]{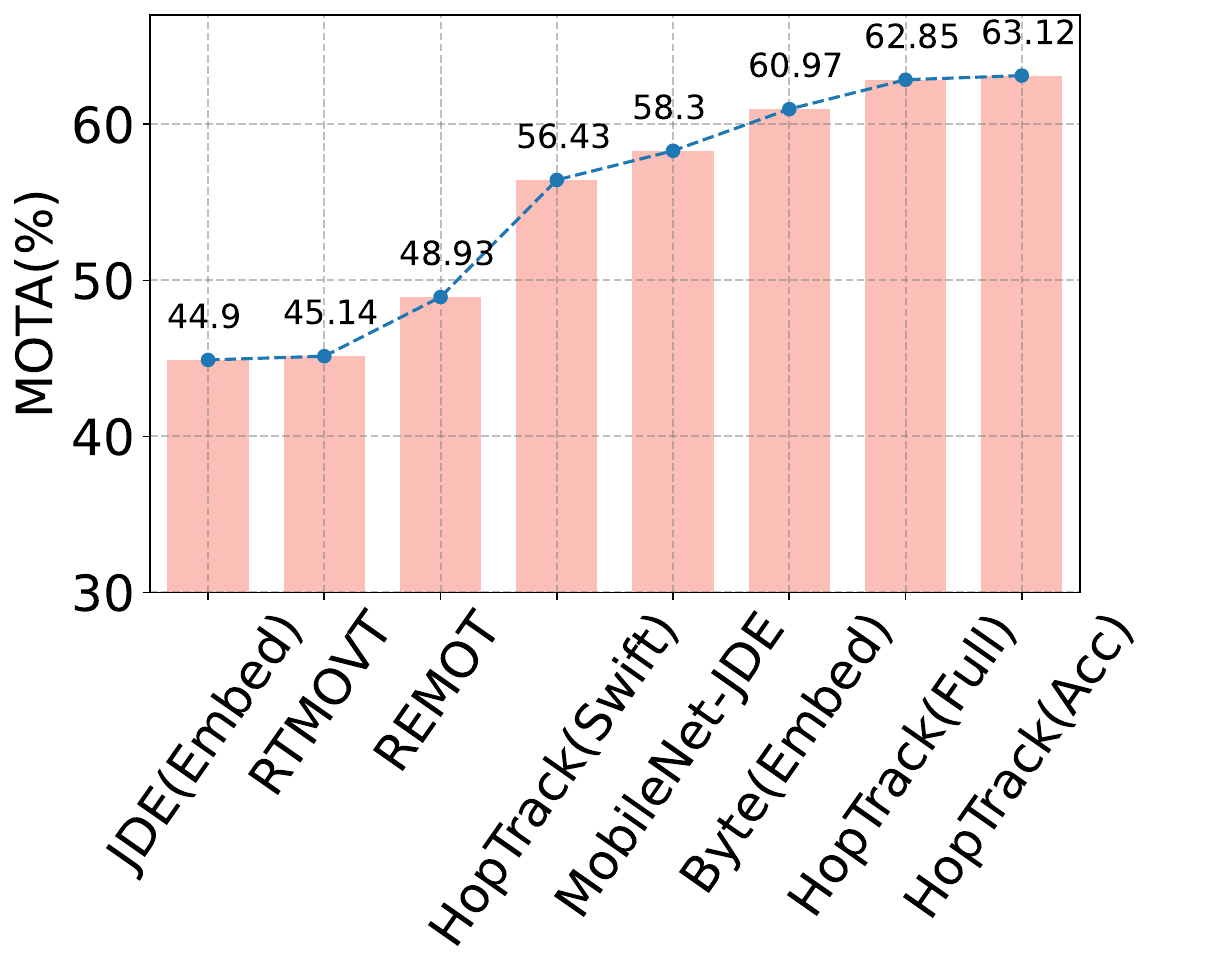}
    \vspace{-7mm}
    \caption{MOTA improvement for MOT16 dataset on embedded\\ devices across 2020 - 2024. }
    \label{fig: mota_imp}
\end{minipage}\hfill
\begin{minipage}[t]{.33\textwidth}
\centering
 \captionsetup{justification=centering}
\includegraphics[width=1.0\columnwidth]{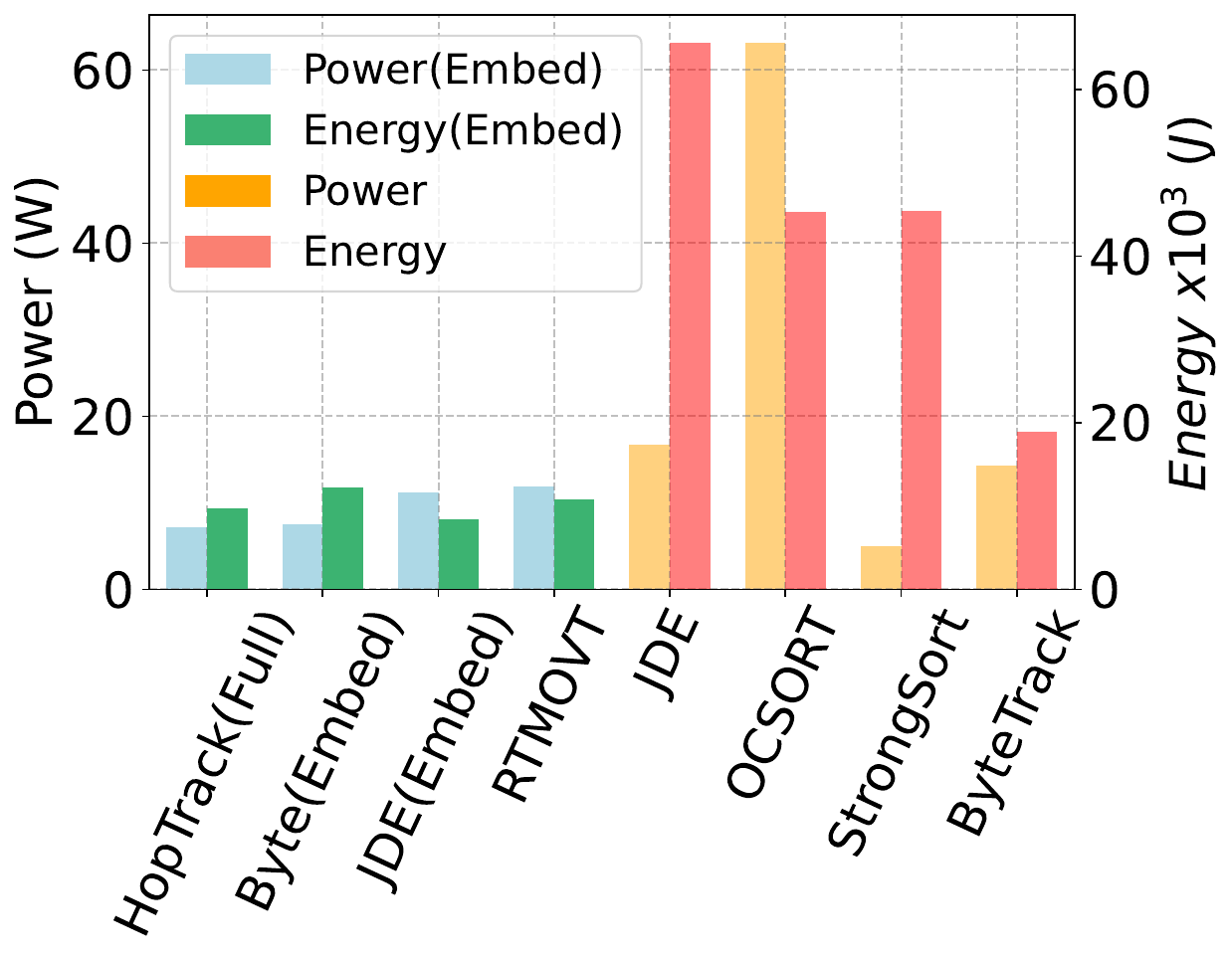}
\vspace{-7mm}
     \caption{\name achieves the least energy consumption and power usage compare to baselines.}
     \label{fig: power_mem}
\end{minipage}
\hfill
\begin{minipage}[t]{.33\textwidth}
\centering
 \captionsetup{justification=centering}
\includegraphics[width=1.0\columnwidth]{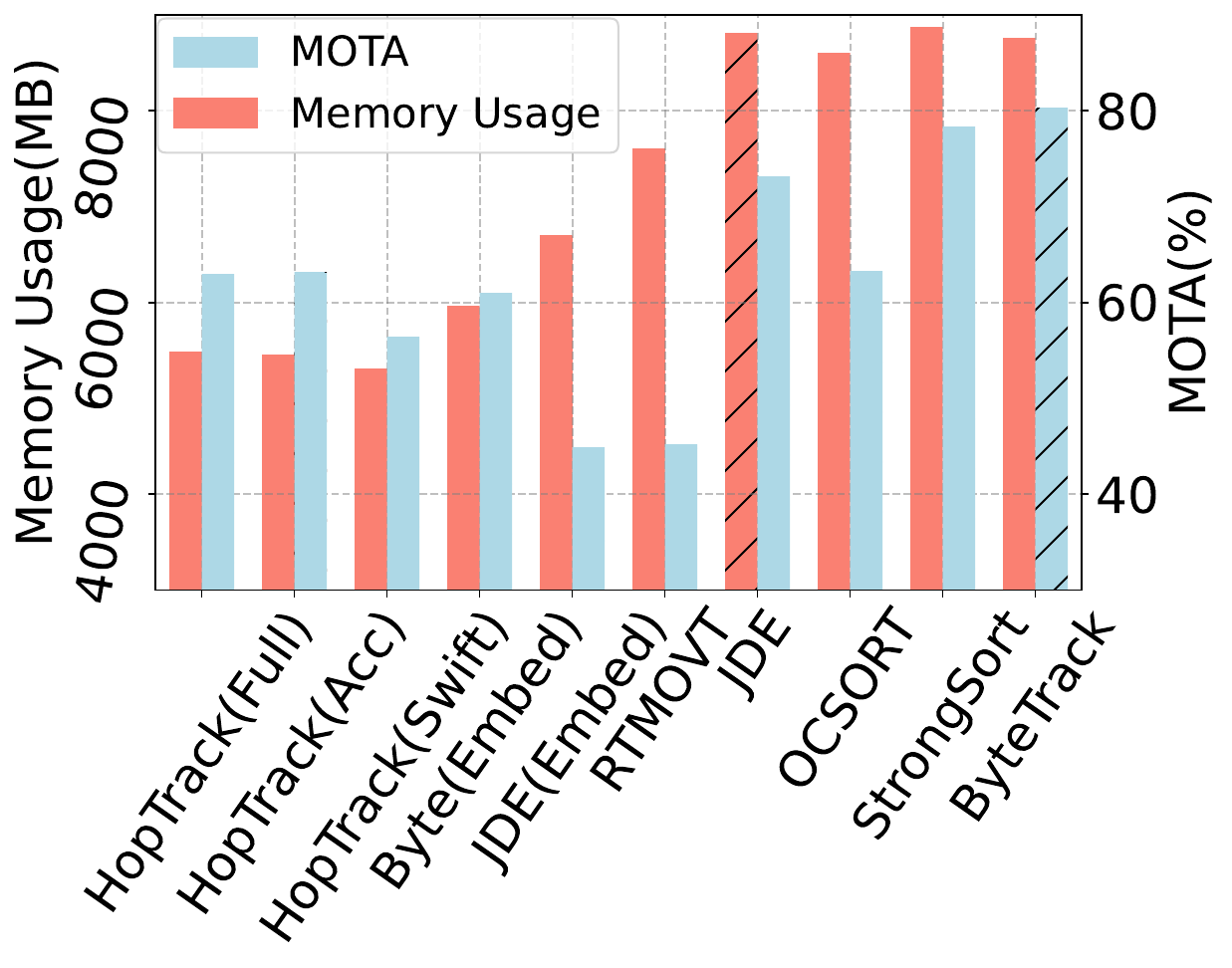}
\vspace{-7mm}
      \caption{
    Memory vs. MOTA. Bars with hatches are methods designed for high-end GPU.}
     \label{fig: mem_mota}
\end{minipage} 

\end{minipage}
\end{figure*}

To delve into the importance of trajectory-based matching, we performed the same test on the MOT16 dataset with trajectory-based matching disabled. We choose the HopTrack(Swift) variant for comparison as it samples at a lower frequency, thus, the object displacement between two detection frames is significant and it adds more difficulty for identity association. In Table~\ref{tbl:hopswift_traj}, all performance metrics are affected when trajectory-based matching is disabled. Specifically, we observed that for MOT16-14 which was filmed in a moving vehicle, trajectory-based matching boosts MOTA, IDF1, and HOTA by 7.12\%, 4.01\%, and 3.91\% respectively and reduces identity switch by 14.5\%. To further demonstrate the effectiveness of trajectory-based matching, we extended our test to video clips from the autonomous driving dataset KITTI~\cite{kitti}. Table ~\ref{tbl:kitti} shows that \name outperforms Byte(Embed) by a large margin.

\begin{table}[h]
  \resizebox{\linewidth}{!}{
  \begin{tabular}{ccccrrr}
  \toprule[.2em]
  \multicolumn{1}{c}{\textbf{Sequence}}      & \multicolumn{1}{c}{\textbf{MOTA\%}}  & \multicolumn{1}{c}{\textbf{IDF1\%}} & \multicolumn{1}{c}{\textbf{HOTA\%}}   & \multicolumn{1}{c}{\textbf{FP}}     & \multicolumn{1}{c}{\textbf{FN}}    & \multicolumn{1}{c}{\textbf{IDSW}} \\
  \midrule
\myrowcolour
\textbf{\name(Full)} & & & & & &\\
 - KITTI-16 & 48.68 & 66.96 & 47.20 & 408   & 438  & 27   \\
 - KITTI-19 & 48.55 & 63.20 & 44.30 & 1100  & 1580 & 69  \\
\myrowcolour
\textbf{BYTE(Embed)} & & & & & &\\
 - KITTI-16 & 36.74 & 59.74 & 40.50 & 456   & 584  & 36   \\
 - KITTI-19 & 37.51 & 55.53 & 39.68 & 1250  & 2021 & 68  \\
\bottomrule
\end{tabular}
}
\caption{Comparison of Hoptrack(Full) and BYTE(Embed) on autonomous driving KITTI dataset.}
\label{tbl:kitti}
\vspace{-5mm}
\end{table}

\begin{table}[h]
  \resizebox{\linewidth}{!}{
  \begin{tabular}{lcccccc}
  \toprule[.2em]
  \multicolumn{1}{c}{\textbf{Scheme}}      & \multicolumn{1}{c}{\textbf{MOTA\%}}  & \multicolumn{1}{c}{\textbf{IDF1\%}} & \multicolumn{1}{c}{\textbf{HOTA\%}}   & \multicolumn{1}{c}{\textbf{FP}($10^{3}$)}     &
  \multicolumn{1}{c}{\textbf{FN}($10^{4}$)}    & \multicolumn{1}{c}{\textbf{IDSW}} \\
  \midrule
HopTrack(Full) & \textbf{45.6} & \textbf{44.5} & \textbf{35.0} & 40.5 & \textbf{23.8} & 2996 \\
\myrowcolour
RTMOVT & 30.0 & 23.0 & 18.6 & 29.1 & 32.4 & 8936 \\
Byte(Embed) & 44.6 & 43.9 & 34.1 & 30.8 & 25.3 & \textbf{2969} \\
\myrowcolour
JDE(Embed) & 34.1 & 27.2 & 21.1 & \textbf{21.1} & 31.2 & 8404 \\
\bottomrule
\end{tabular}%
}
\caption{MOT20 tracking results comparison.}
\label{tbl:mot20_comparison}
\end{table}

We also performed experiment using \name on the MOT17, the results are available in our anonymized repository, where similar results are observed. The detailed performance comparison among \name and baselines on the MOT20 dataset is documented in Table \ref{tbl:mot20_comparison}. Compared with MOT16 and MOT17, which have an average object density of 30.8 objects/frame with a peak of 69.7 objects/frame, the MOT20 dataset is much more challenging with an average object density of 170.9 objects/frame and a peak of 205.9 objects/frame. Such high density accentuates the tracking difficulties as more tracks can interfere with each other and even a short detection gap could lead to identity association trouble. Unsurprisingly, no prior framework on embedded devices is capable of real-time MOT on the MOT20 dataset and the performance of \name degrades on the MOT20 dataset. Such performance degradation compared with the MOT16/17 test sequence is observed for other baselines, as well as the state-of-the-art trackers running on high-end GPUs such as ByteTrack~\cite{zhang2022bytetrack}, JDE~\cite{jde}, etc. We emphasize that to the best of our knowledge, \name is the first multi-object tracking framework designed for embedded devices that officially reports the accuracy metrics on the MOT20 test dataset, which sets a new state-of-the-art for this category. 

\subsubsection{\textbf{Tracking Latency}}~
A comparison of average tracking time with respect to the object density is shown in Figure~\ref{fig: tracking_time}. We observed that there exists a linear relationship between the tracking time and the object density, but the slopes and the offsets of the different solutions differ quite significantly. 
\name(Full) and our modified baseline Byte(Embed) set the best process rate at 13.94 fps and 17.74 fps respectively, which are calculated as the average inference time and tracking time across the entire MOT20 dataset. When the object density increases, the tracking time increases sharper for some solutions than others. The smallest increase gradients are for \name(Full) and Byte(Embed). It is worth noting that due to the object density increase in the MOT20 dataset, none of the existing methods, including \name, can perform real-time tracking on the embedded device. 

\subsubsection{\textbf{Memory, Power and Energy Consumption}}~
Given that \name is designed for embedded devices, evaluating its power and energy efficiency as well as other precious on-node resources is crucial. To this end, we leveraged the \texttt{tegrastats} API of the Jetson platform to monitor key device metrics throughout the framework's runtime, including CPU and GPU utilization, memory consumption, and power usage. The figures presented in this section reflect the average energy, power (including both CPU and GPU), and memory consumption observed over the entire MOT16 dataset.

Figure~\ref{fig: mem_mota} shows the memory usage vs tracking accuracy for \name and baselines. \name consumes the least memory while achieving the highest tracking accuracy compared with all baselines proposed for embedded devices. Figure \ref{fig: power_mem} shows the combined power consumption of both CPU and GPU. Energy consumption is measured as the product of the average power over the entire MOT16 dataset (all video sequences) and the total execution time.

Compared with tracking methods designed for high-end GPUs, frameworks designed for embedded devices consume far less power and energy. While StrongSort shows the lowest average power consumption, its extended execution time places it second in energy consumption. Moreover, we can also observe that joint detection and embedding based methods such as JDE, JDE(Embed), and RTMOVT consume more power due to the extra embedding feature inference.

\subsubsection{\textbf{Performance under Resource Contention}}~ \label{subsubsec:rescontention}
%(3/15/24) XL%We have observed a trend in which large infrastructure providers~\cite{sagemaker} are deploying edge devices with GPU capabilities across the edge networks. These devices are often shared publicly and used on-demand, which can lead to potential resource contention issues.
To quantitatively evaluate \name's performance under resource contention at the network's edge, we conducted separate CPU and GPU stress tests. For the CPU stress test, we utilized the \textbf{\texttt{stress}} workload generator tool to ramp up the workload on each CPU core. For the GPU load generation, we used our developed customized tool capable of generating specified GPU loads with an error rate of less than 5\% (included in our open source package referenced in the abstract). Test results are presented in Figures~\ref{fig: gpu_stress} and~\ref{fig: cpu_stress}.
\begin{figure*}
\begin{minipage}[t]{.32\textwidth}
\centering
\includegraphics[width=1.0\columnwidth]{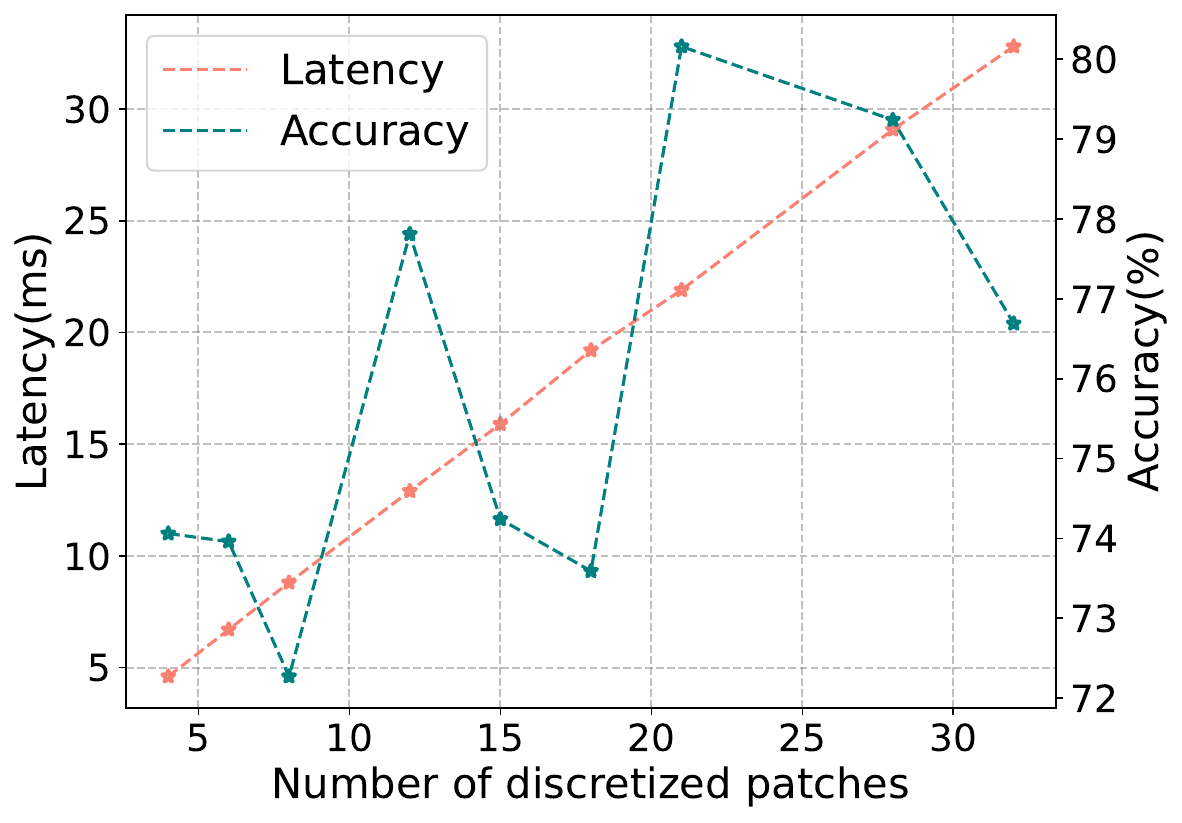}
\vspace{-7 mm}
\caption[]{Latency and accuracy trade-off as the number of discretized patches increases. The trend line is for increasing accuracy and increased latency, though accuracy has discontinuities.}
\label{fig:dis_lat}
\end{minipage}\hfill
\begin{minipage}[t]{.28\textwidth}
\centering
\includegraphics[width=\columnwidth]{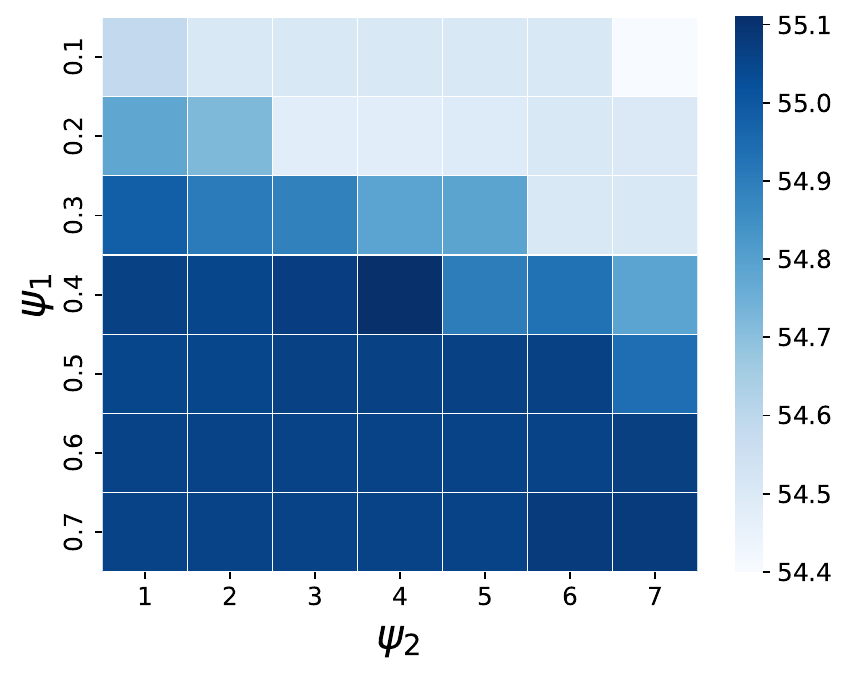}
\vspace{-7 mm}
     \caption{Sensitivity of accuracy to choices of $\psi_{1}$ and $\psi_{2}$ used during the discretized static matching. It is relatively insensitive in a wide region.}
     %\mustafa{Explain what do you mean by stress in the caption}}
     %(03/14/23) XL: addressed the above comments
    \label{fig:sensitivity1}
\end{minipage}\hfill
\begin{minipage}[t]{.28\textwidth}
\centering
\includegraphics[width=1.04\columnwidth]{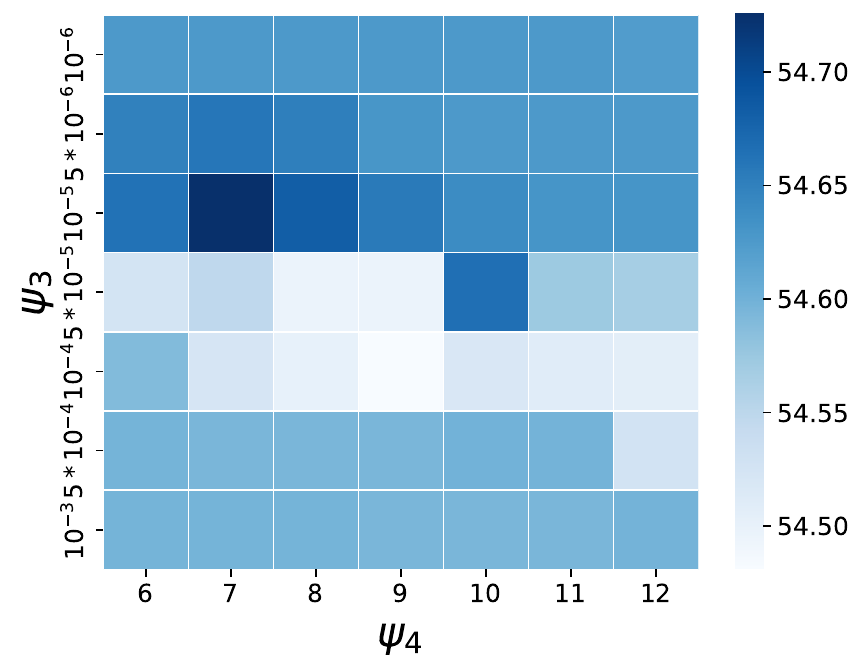}
\vspace{-7 mm}
     \caption{Sensitivity of accuracy to choices of $\psi_{3}$ and $\psi_{4}$ used  during the discretized dynamic matching. It is relatively insensitive in a wide region.} 
    \label{fig:sensitivity2}
     %\kwang{We'll need to reposition the legend such that it no longer meets lines.}}
     % XL(03/17/23): fixed the legend position issue
\end{minipage}\hfill
% \end{minipage}
\vspace{-3mm}
\end{figure*}
We observed increasing GPU contention extended inference time from 75 ms to more than 600 ms. However, the tracking algorithm, executed on the CPU, remained unaffected by GPU contention. Tracking time saw no significant impact until the last free CPU core is stressed, when the tracking time increases from 21 ms to 38 ms. This aligns with expectation, given that for this run, we configured our tracking algorithm to utilize only one CPU core. We do notice a slightly increased inference time when the CPU cores are stressed, this may be attributed to CPU-related post-processing and frame loading operations.
\footnotetext{bars with hatches are methods designed for high-end GPU}

\subsubsection{\textbf{Detector Agnostic Design}}~ \label{subsec:detector_agnostic}
\name is designed to be detector agnostic. It can be integrated with a multitude of detectors to perform multi-object tracking. This is practically important as object detection is a fast-moving field with regular improvements. Besides YOLOX, we also integrated \name with the state-of-the-art detector YOLOv7~\cite{yolov7} and its edge GPU version YOLOv7-tiny.
\begin{table}[th]

  \resizebox{\linewidth}{!}{
  \begin{tabular}{lccccc}
  \toprule[.2em]
  \multicolumn{1}{c}{\textbf{Detector}}      & \multicolumn{1}{c}{\textbf{MOTA\%}}  & \multicolumn{1}{c}{\textbf{IDF1\%}}  & \multicolumn{1}{c}{\textbf{FP}($10^{3}$)}     &
  \multicolumn{1}{c}{\textbf{FN}($10^{4}$)}    & \multicolumn{1}{c}{\textbf{IDSW}} \\
  \midrule
HopTrack(Full)+YOLOX-S & \textbf{62.91} & 60.83 & 19.1 & \textbf{46.2} & 2278 \\
\myrowcolour
HopTrack(Full)+YOLOv7-tiny & 61.45 & 61.57 & 17.2 & 50.5 & 2541 \\

HopTrack(Full)+YOLOv7 & 56.6 & 59.4 & 26.3 & 50.5 & 2370 \\
\myrowcolour
Byte(Embed)+YOLOv7-tiny & 60.15 & \textbf{62.4} & \textbf{16.7} & 53.9 & \textbf{2050} \\
Byte(Embed)+YOLOv7 & 55.7 & 60.07 & 26.1 & 51.4 & 3150 \\
\bottomrule
\end{tabular}%
}

\caption{\name outperforms ByteTrack under YOLOv7 as well as YOLOX detectors on the MOT16 test dataset. This shows the ability of \name to pair with different generations of detectors.}
\label{tbl:yolov7}
\vspace{-5mm}
\end{table}

YOLOv7 has shown significant inference speed improvement, with YOLOv7-tiny running at 25-35 ms per frame and YOLOv7 at 60-75 ms per frame on AGX. However, the original YOLOv7 only reported the accuracy on the COCO ~\cite{COCO} dataset, in which less than 1\% of images contain more than 6 objects and cannot reflect the MOT scenario. We fine-tuned the YOLOv7 and YOLOv7-tiny following the same procedure as we did with YOLOX-S and integrated it with BYTE(Embed) and \name. The test result is shown in Table \ref{tbl:yolov7}. We observe that \name + YOLOv7 outperforms BYTE(Embed) + YOLOv7, but the performance still falls short compared with \name integrated with YOLOX-S. A potential explanation is that anchor-based detectors such as YOLOv7 are less generalizable compared with anchor free detectors like YOLOX, which leads to inferior performance in crowded scenes.
% SB (11/30/23): What is meant by "less generalizable"? Less generalizable to different kinds of scenes?

\subsubsection{\textbf{Hyper-parameter Sensitivity Test}}~
\label{sec:sensitivity}
MOT16 and MOT20 datasets cover a great variety of scenarios, including videos taken from an elevated view, from handheld mobile devices, and from a vehicle in motion. The current empirically determined configuration should be able to adapt to this variety of scenarios. We perform sensitivity studies for the tunable parameters in discretized static and dynamic matching. Figures \ref{fig:dis_lat}, \ref{fig:sensitivity1}, and \ref{fig:sensitivity2} show the relationship between the latency and the number of discretized cells, the effect of the matching thresholds $\psi_{1}-\psi_{4}$ (Section~\ref{subsec: image_quantization}) on the MOTA respectively\footnotemark{}. \xiang{We use the optimal configuration from these results to evaluate other datasets, with the aim of generalizing the framework’s performance rather than fine-tuning for a specific dataset.} We see from Figure~\ref{fig:dis_lat} that there exists a trade-off between computation latency and the matching accuracy as the number of discretized cells increases. The fluctuation of MOTA is due to the different effects of increasing the number of patches in rows versus in columns. 
% SB (11/30/23): Verify
From Figures~\ref{fig:sensitivity1} and \ref{fig:sensitivity2}, we see there are large operating regions for the parameter values to provide comparable performances. For the rare scenarios where the current configuration is poor, one can easily tune those configurations accordingly.

\footnotetext{Completely optional material with detailed sensitivity studies in static and dynamic matching is available in our anonymous repository.}

\section{Related Work} \label{sec:related_work}

\noindent \textbf{Object detection on embedded devices.} Object detection 
aims to detect and localize semantic objects within images or video streams. It can be categorized into two-stage detectors and one-stage detectors. Two-stage detectors, such as R-CNN~\cite{girshick2014rich}, Fast R-CNN~\cite{girshick2015fast}, Faster R-CNN~\cite{ren2015faster} perform proposal generation using Selective Search~\cite{uijlings2013selective} or Region Proposal Network (RPN)~\cite{ren2015faster}, followed by proposal classification in the sparse set of candidate object locations. In contrast, one-stage detectors, such as RetinaNet\cite{lin2017focal}, YOLO series~\cite{redmon2016you, redmon2018yolov3, ge2021yolox, yolov7}, and SSD~\cite{liu2016ssd}, perform both region proposal and classification in a single step or skip the proposal step altogether, making them faster but potentially less accurate. 
Achieving real-time object detection on embedded devices is challenging. Frameworks such as Glimpse~\cite{naderiparizi2017glimpse} and Glimpse~\cite{chen2015glimpse} intelligently select keyframes for offloading to the cloud. In contrast, ApproxDet~\cite{xu2020approxdet} and Chanakya~\cite{ghosh2023chanakya} aim to achieve accuracy-latency Pareto decisions by dynamically generating perception configurations, such as input resolution and number of proposals.

\noindent \textbf{MOT}. The objective of MOT is to associate detections across frames. However, several factors, such as motion variations and cluttered backgrounds, make this task challenging. 
Several MOT methods have been proposed to address these challenges.
SORT~\cite{sort} achieves high efficiency while maintaining simplicity by combining the Kalman filter for motion estimation and the Hungarian algorithm~\cite{kuhn1955hungarian} for data association, yet it lacks robustness in complex scenes that involve variant appearance features.
DeepSort~\cite{deepsort1} overcomes this issue by integrating visual descriptors extracted by a pre-trained deep CNN for similarity measurement, while StrongSort~\cite{du2022strongsort} improves it further by using an advanced detector and appearance feature extractors.
ByteTrack ~\cite{zhang2022bytetrack} conducts a two-stage association by separating detection boxes into high and low confidence score ones, different from methods ~\cite{zhang2021fairmot, pang2021quasi,jde} that only keep the high confidence detection boxes.
 
% \mike{Mar 08, 2023}
All aforementioned methods fall under the category of Separate Detection and Embedding (SDE), which includes a detection model for target localization and a feature embedding model to extract re-identification information for data association. Joint Detection and Embedding (JDE)~\cite{jde,zhang2021fairmot,pang2021quasi} is another paradigm for MOT, where object detection and tracking models are jointly optimized. For example, Tracktor~\cite{bergmann2019tracking} directly adopts an object detector for tracking, while RetinaTrack~\cite{lu2020retinatrack} modifies the Post-FPN subnetwork of RetinaNet~\cite{lin2017focal} to capture instance-level embeddings. 
Our work falls under neither category as we do not use an embedding model for feature extraction. 

\noindent\textbf{MOT on embedded device.} To address the challenges in Section~\ref{sec:problem_statement} for embedded devices, RTMOVT~\cite{rtmot} combined JDE-modified YOLOv3 with a Kalman filter and KCF tracker~\cite{henriques2014high} to boost the association accuracy. REMOT~\cite{remot} enhances the feature embedding model with an angular triplet loss to increase the re-identification accuracy. A very recent paper~\cite{li2023multi} claims to do multi-object tracking on IoT devices through novel object-aware embedding, which enables them to achieve accurate, lightweight association. However, their evaluation is largely done on desktop GPUs (GTX 1080Ti).
\section{Discussion}\label{sec:discussion}

An aspect where \name can be an enabler is where queries selectively look for a specific object(s) across frames, such as, in surveillance situations where the query may be for a specific vehicle~\cite{xu2022rev}. Then our matching technique can be used to quickly eliminate all the objects that are not being queried and perform matches for the object of interest.

Another emerging application that our work helps toward is a real-time, multi-camera object tracking system for smart cities~\cite{zhang2018person, yang2018distributed}. This system involves deploying multiple cameras throughout the area of interest, with each camera feeding video streams to an embedded device for real-time processing. Since our work reduces the need for heavy-duty GPU processing in favor of lightweight operation (tracking), it would be possible to deploy this on relatively wimpy GPU nodes at scale. 

Our work represented here is a natural progression in the highly active area of adapting ML models and inferencing for smaller devices. Within this, our work addresses one important class of streaming video analytics, namely multi-object tracking. Ours is fundamentally a systems problem as the protocol has to meet the resource constraints of embedded devices, e.g., we consciously design to stay under the memory capacity of even an entry level mobile GPU, the Jetson TX2 (8 GB).
This represents a crucial step to enabling continuous vision even with complex videos, as it enables the relatively lightweight tracking to occur more frequently and reduce the frequency of invoking the more expensive detection algorithm. Adaptivity is an important dimension in this class of algorithms to deal with changing video content characteristics (which we account for) and changing availability of network resources due to contention or variability (which we do not yet). 

We recognize the considerable effort invested in developing light-weight detection models within the computer vision field to enable faster and more accurate inference on edge GPUs (i.e. Yolov7-tiny). However, our findings in Section~\ref{subsec:detector_agnostic} show that despite employing the state-of-the-art edge GPU detector capable of near real-time inference on AGX Xavier, the MOT problem remains challenging. \xiang{Incorporating advanced detection models can further enhance HopTrack’s performance. Similarly, advancements in embedded devices can improve detection inference time and enable a higher frame rate for detection.}
\section{Conclusion} \label{sec:conclusion}
We presented \name, a framework for real-time multi-object tracking on resource-constrained embedded devices.
\name achieves accurate tracking for fast-moving objects through novel trajectory-based data association and discretized static and dynamic matching. To handle complex scenes and to enhance robustness against occlusion, \name incorporates innovative content-aware dynamic sampling for improved object status estimation during occlusion. 
\name surpasses existing methods (i.e., Byte(Embed) and MobileNet-JDE) by enhancing processing rate (frames per second) and accuracy.
Our experiments show that \name achieves state-of-the-art performance on NVIDIA Jetson AGX with 63.12\% MOTA and 39.29 fps at low resource consumption on the MOT16 test dataset.
Further, \name's detector-agnostic nature facilitates seamless integration with light-weight detectors, making it an ideal tracker for real-world applications that have limited computing resources and stringent latency requirements.
\balance
\bibliographystyle{IEEEtran}
\bibliography{reference}

\end{document}